\documentclass[final]{cvpr}

\usepackage{times}
\usepackage{epsfig}
\usepackage{graphicx}
\usepackage{amsmath}
\usepackage{amssymb}
\usepackage{xcolor}
\usepackage{array}
\usepackage{booktabs}
\usepackage{enumitem}
\usepackage{subfig}
\usepackage{float}
\usepackage{blindtext}
\usepackage{caption}

\usepackage[pagebackref=true,breaklinks=true,colorlinks,bookmarks=false]{hyperref}

\def\cvprPaperID{****} 
\def\confYear{CVPR 2021}

\newcommand\blfootnote[1]{%
  \begingroup
  \renewcommand\thefootnote{}\footnote{#1}%
  \addtocounter{footnote}{-1}%
  \endgroup
}

\newcommand{\density}{\sigma}
\newcommand{\radiance}{\mathbf{c}}
\newcommand{\mlp}{\mathcal{F}_{\theta}}
\newcommand{\ray}{\mathbf{r}}

\newcommand{\R}{\mathbb{R}}

\newcommand{\SE}[1]{\mathbf{SE}{(#1)}}

\newcommand{\mask}[1]{M}

\newcommand{\se}[1]{\mathfrak{se}{(#1)}}

\newcommand{\volume}{\mathop{\ooalign{\hfil$V$\hfil\cr\kern0.08em--\hfil\cr}}\nolimits}

\begin{document}

\title{STaR: Self-supervised Tracking and Reconstruction of Rigid Objects in Motion with Neural Rendering}

\author{
Wentao Yuan*\\
University of Washington
\and
Zhaoyang Lv \qquad
Tanner Schmidt \qquad
Steven Lovegrove\\
Facebook Reality Labs Research
\and
\url{https://wentaoyuan.github.io/star}
}

\makeatletter
\let\@oldmaketitle\@maketitle
\renewcommand{\@maketitle}{\@oldmaketitle
    \centering
    \includegraphics[width=0.9\linewidth]{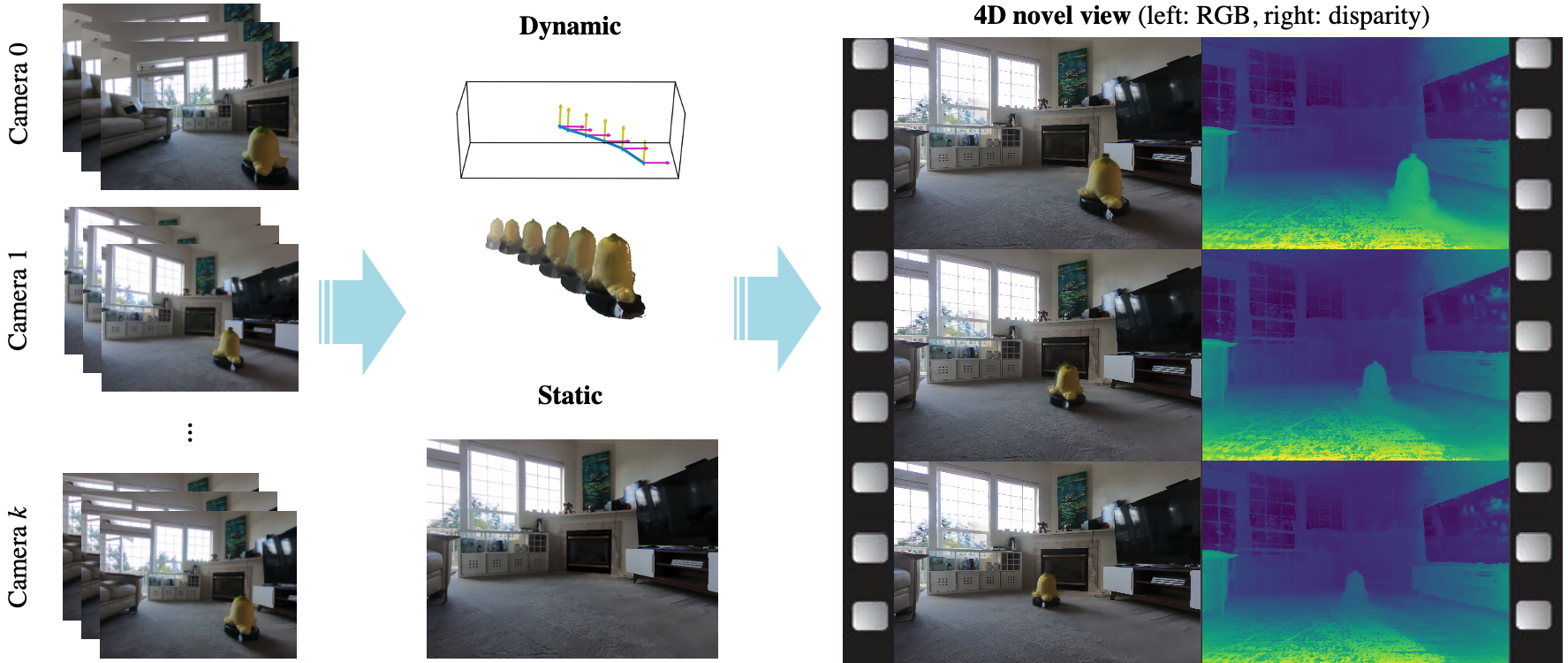}
    \captionof{figure}{\small \textbf{An overview of our method.} Given multi-view RGB videos of a dynamic scene, STaR learns a decoupled 3D representation of the static and dynamic scene components without any human annotation, which allows it to synthesize the scene from new viewpoints at new time photorealisticly, or even animate the scene with novel trajectories.}
    \label{fig:teaser}
    \vspace{3mm}
}

\makeatletter
\renewcommand{\paragraph}{%
  \@startsection{paragraph}{4}%
  {\z@}{1ex \@minus .2ex}{-1em}%
  {\normalfont\normalsize\bfseries}%
}
\makeatother

\maketitle
\blfootnote{*Work done while the author was an intern at FRL Research.}

\begin{abstract}
We present STaR, a novel method that performs Self-supervised Tracking and Reconstruction of dynamic scenes with rigid motion from multi-view RGB videos without any manual annotation. Recent work has shown that neural networks are surprisingly effective at the task of compressing many views of a scene into a learned function which maps from a viewing ray to an observed radiance value via volume rendering. Unfortunately, these methods lose all their predictive power once any object in the scene has moved. In this work, we explicitly model rigid motion of objects in the context of neural representations of radiance fields. We show that without any additional human specified supervision, we can reconstruct a dynamic scene with a single rigid object in motion by simultaneously decomposing it into its two constituent parts and encoding each with its own neural representation. We achieve this by jointly optimizing the parameters of two neural radiance fields and a set of rigid poses which align the two fields at each frame. On both synthetic and real world datasets, we demonstrate that our method can render photorealistic novel views, where novelty is measured on both spatial and temporal axes. Our factored representation furthermore enables animation of unseen object motion.
\end{abstract}

\vspace{-1.5em}
\section{Introduction}

Recent years have seen an explosion of novel scene representations which leverage multi-layer perceptrons (MLPs) as an implicit representation to encode spatially-varying properties of a scene. While these implicit representations are optimized via stochastic gradient descent, they are not really ``learning" in the traditional sense. 
Instead, they exploit MLPs as a compressed representation of scene content. They can act as a substitute for traditional representations (e.g. explicit volumetric grids), but can adaptively distribute its limited capacity over the scene to enable high-fidelity scene representation.
Representative examples include DeepSDF \cite{park:etal:cvpr19}, Scene Representation Networks \cite{sitzmann:etal:NeurIPs19} and Neural Radiance Fields (NeRF) \cite{mildenhall:etal:arXiv20}.

Among all of these representations, NeRF \cite{mildenhall:etal:arXiv20} and its variants \cite{liu:etal:neuralips20, zhang:etal:arXiv2020, martin:etal:arXiv20} show enormous potential in their ability to photorealistically reconstruct a scene from only a sparse set of images. However, these works assume the scene is static, or at least that the dynamic content of the scene is uninteresting and can be discarded as in Martin-Brualla et al. \cite{martin:etal:arXiv20}. 
When objects in a scene move, these methods can no longer correctly render novel views. 
It is possible to represent time-varying scenes by dedicating one NeRF volume per frame or by extending the input to four dimensions by including time \cite{sitzmann:etal:arXiv20, tancik:etal:neuralips20}. The former is unnecessarily expensive, and neither can be used to render the object in novel poses (or remove it entirely) because they have no object-level understanding of the scene. 

In this work, we aim to learn an interpretable and editable representation of a dynamic scene by simply observing an object in motion from multiple viewpoints. As an initial effort to tackle this challenge, we start from a simplified setting that assumes the scene contains only one moving object and the motion is fully rigid.
We accomplish this by rendering a compositional neural radiance field with density and radiance defined via a composition of a static and a dynamic neural radiance field. 
Under this model, the only way all observations in a video can accurately be predicted is by segmenting the scene into the two volumes and correctly estimating the pose of the object in each frame. 

To achieve this goal, our paper provides two main technical contributions. First, we present the first self-supervised neural rendering based representation that can simultaneously reconstruct a rigid moving scene as well as its background from videos only. Our approach enables photorealistic spatial-temporal novel view rendering as well as novel scene animation. Second, we present an optimization scheme that can effectively tackle the ambiguity and local optima during training. 

Our experiments show that it is possible to recover the segmentation of static and dynamic contents as well as motion trajectory without any supervision other than multi-view RGB observations.
Compared to NeRF and its extensions, our approach achieves more photorealistic reconstruction in complex synthetic as well as real world scenes.
Further, our factorized representation can be edited to position the object in novel locations never observed during training, which no existing method can achieve without 3D ground truth or supervision. 


\section{Related Work}

Our work is inspired and related to the recent progress in learning-based representations, and particular in novel differentiable rendering from images, e.g. NeRF \cite{mildenhall:etal:arXiv20} and its variants \cite{martin:etal:arXiv20, zhang:etal:arXiv2020}. However, no prior work has successfully demonstrated photorealistic reconstruction and understanding of dynamic scenes using only real-world natural images.
We believe we are the first to achieve self-supervised tracking and reconstruction of dynamic (rigidly) moving scenes using neural rendering. 

\paragraph{Neural implicit representation}
Recently deep implicit representations using MLPs have demonstrated a promising ability to learn high quality geometry and appearance representations. Existing works use MLPs to represent the scene as an implicit function which maps from scene coordinates to scene properties, e.g. signed distance functions (SDFs) \cite{park:etal:cvpr19}, occupancy \cite{mescheder:etal:cvpr19}, occupancy flow \cite{niemeyer:etal:cvpr19}, volumetric density and radiance \cite{mildenhall:etal:arXiv20}, or an implicit feature \cite{sitzmann:etal:NeurIPs19} which can be further decoded into pixel appearance. Several approaches show high quality reconstruction results by pairing such implicit representations with differentiable ray marching and some manual input such as object masks \cite{niemeyer:etal:cvpr20} \cite{yariv:etal:neuralips20}. Sitzmann \etal \cite{sitzmann:etal:arXiv20} and Tancik \etal \cite{tancik:etal:neuralips20} shows implicit representation can compress videos, but has yet demonstrated its ability to reconstruct dynamic 3D scenes.

NeRF \cite{mildenhall:etal:arXiv20} demonstrated that a continuous implicit volume rendering function can be learned from only sparse sets of images, achieving state-of-the-art novel view synthesis results without relying on any manually specified inputs. Recent work has extended this representation to be trained with web images under different lighting conditions \cite{martin:etal:arXiv20}, with arbitrary viewpoints in a real-world large-scale scene \cite{zhang:etal:arXiv2020}, with hierarchical spatial latent priors \cite{liu:etal:neuralips20}, or with a latent embedding that can support generative modeling \cite{schwarz:etal:neuralips20}. However, existing works all assume the input scene coordinate is quasi time-invariant while we assume the scene is rigidly moving. Among all of these work, only NeRF-W \cite{martin:etal:arXiv20} consider the real-world that contains dynamic information. They incorporated a latent embedding in addition to the scene coordinate to model the photometric variation across web images and treat the moving scene as transient objects. This method discards these objects and only reconstructs the background scene, while we can simultaneously reconstruct dynamic objects and recover their trajectory.

\paragraph{Dynamic scene novel view synthesis}
Recently a few systems demonstrate novel view synthesis in dynamic scenes using multi-view videos, with a neural image-based rendering representation \cite{Bemana:etal:siggraphasia20}, neural volume rendering \cite{lombardi:etal:tog19}, or a multi-spherical image representation \cite{broxton:etal:siggraph20}. However, these approaches can only support dynamic video playback; they cannot understand scene dynamics, and therefore cannot be used for interactive animation. Yoon \etal \cite{yoon:etal:cvpr2020} propose a method that blends multi-view depth estimation in background regions, learned monocular depth estimation in the foreground, and a convolutional network. They demonstrated novel view synthesis as well as animation. However, this system requires precise manual specification of the foreground mask as part of training and rendering. This is a significant drawback, especially when it comes to video input. In contrast, we demonstrate that our approach can achieve simultaneous reconstruction of background and foreground completely self-supervised and demonstrate that it achieves high-quality reconstruction using real-world videos. 

\paragraph{Self-supervised learning in dynamic scenes}
There is also a surge in self-supervised learning approaches using videos, particularly in the autonomous driving domain, including 3D object detection \cite{beker:etal:eccv20}, joint camera poses and depth estimation \cite{godard:etal:cvpr19,chen:etal:iccv19}, with motion segmentation \cite{ranjan:etal:cvpr19,luo:etal:pami19} and scene flow \cite{hur:stefan:cvpr20}. Due to the sensor limitation on autonomous driving domain, these approaches focus on monocular videos or narrow baseline stereo videos where the change in viewpoint is small. As a result, these methods use 2D or 2.5D representations which limits their ability to render or animate the scene in a photorealistic fashion. In contrast, our method builds a implicit 3D representation that can be photorealistically rendered and animated. We demonstrate applications of our model on natural wide-baseline multi-view videos, which shows promise for applications in virtual and augmented reality.

\section{Method} \label{sec:method}

\begin{figure*}[t]
    \centering
    \includegraphics[width=0.75\linewidth]{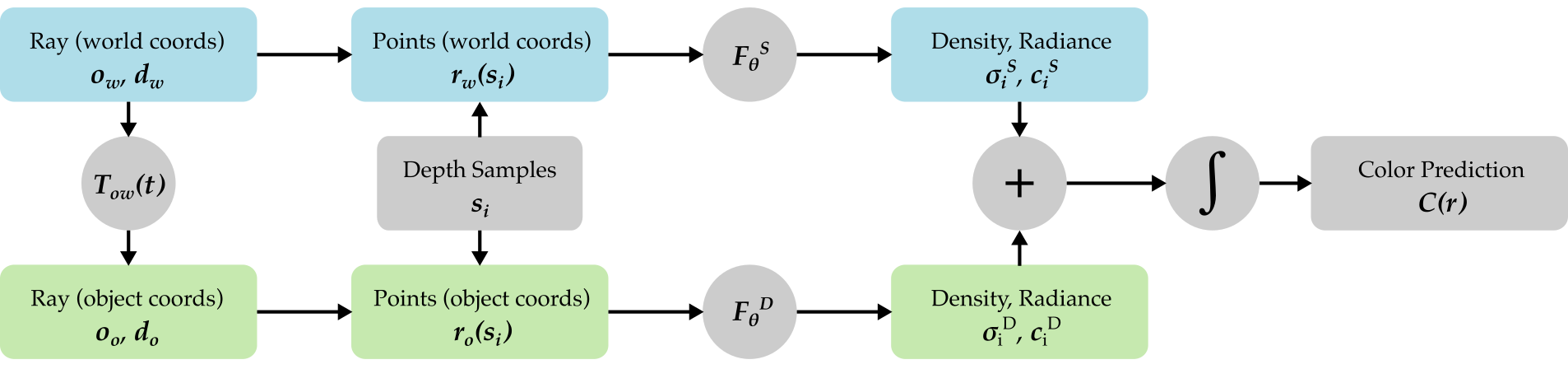}
    \caption{\small Illustration of the architecture of STaR. In order to cast a ray through the composed dynamic volume, we sample points on the original ray as well as the equivalent ray in the object's canonical frame of reference. Samples are passed to the corresponding neural networks, and resulting radiance and density values (now coordinate-free) are combined by addition.}
    \label{fig:arch}
    \vspace{-1em}
\end{figure*}

We introduce STaR, a differentiable rendering-based framework for self-supervised tracking and reconstruction of rigidly moving objects and scenes. 
Given only multi-view passive video observations of an unknown object which rigidly moves in an unseen environment, STaR can simultaneously reconstruct a 3D model of the object (including both geometry and appearance) and track its 6DoF motion relative to a canonical frame without any human annotation. 
STaR can enable high-quality novel viewpoint rendering of both static and dynamic components of the original scene independently at any time, and can also be animated and re-rendered with a novel object trajectory.

We will first describe the 3D representation of STaR, which consists of a static and a rigidly-moving NeRF model. Next, we elaborate how to optimize STaR over a multi-view RGB video, specifically including how to jointly optimize rigid motion optimization over the Lie group of 3D transforms in the context of neural radiance fields, a regularized loss function, and an online training scheme that can work with videos of arbitrary length.

\begin{figure}[h]
\centering
\subfloat[Scene at $t_0$]{
    \includegraphics[width=0.3\linewidth]{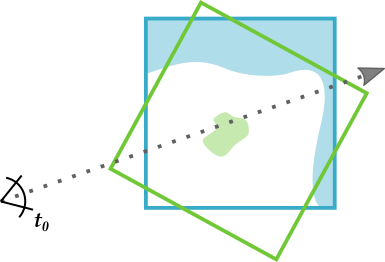}
    \label{fig:t0}
}
\hspace{3mm}
\subfloat[Scene at $t_1$]{
    \includegraphics[width=0.3\linewidth]{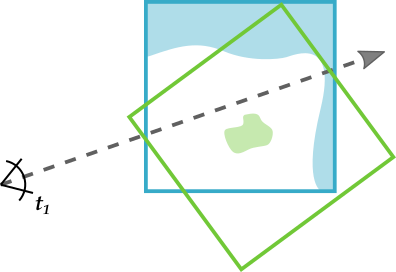}
}
\vspace{-3mm}
\subfloat[Static volume]{
    \includegraphics[width=0.3\linewidth]{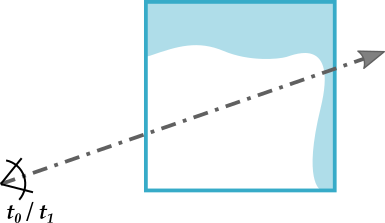}
}
\hspace{3mm}
\subfloat[Dynamic volume]{
    \includegraphics[width=0.3\linewidth]{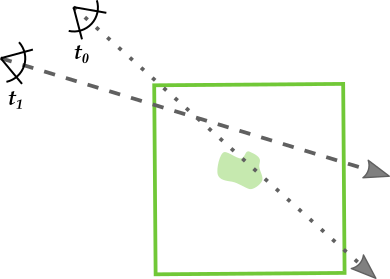}
}
\caption{\small A visualization of the volumetric composition of the two volumes. (a) and (b) show a 2D scene observed at two different time $t_0$ and $t_1$ from a static viewpoint. To render the images observed at $t_0$ and $t_1$, we would cast the same ray through the static volume (shown in (c)), but two different rays through the dynamic volume (shown in (d)), which are transformed copies of the static ray based on the dynamic object's poses at $t_0$ and $t_1$.
}
\vspace{-1.5em}
\end{figure}

\subsection{STaR as Dynamic Neural Radiance Fields}

\paragraph{Preliminaries} NeRF \cite{mildenhall:etal:arXiv20} represents the scene as a continuous function $\mlp$ using a MLP with parameters $\theta$
\begin{align}
    \mlp: \mathbf{r}(s), \mathbf{d} \rightarrow \density, \radiance
\end{align}
which maps a 3D scene coordinate $\mathbf{r}(s)\in \R^3$ and view direction $\mathbf{d} \in S^2$ to volumetric density $\density \in \R$ and radiance $\radiance \in \R^{3}$, where $\mathbf{r}(s)=\mathbf{o}+s\mathbf{d}$ is the point on the ray from camera origin $\mathbf{o}$ with a depth of $s$. The pixel color can be obtained via an integral of accumulated radiance along the ray.
We can numerically estimate the integral using quadrature:
\begin{align} 
    & \hat{C}(\mathbf{r}) = \sum_{i=1}^NT_i\alpha_i\mathbf{c}_i \label{eq:static_radiance}~, \\
    \text{where } & T_i = \exp\left(-\sum_{j=1}^{i-1}\sigma_j(s_{j+1}-s_j)\right) \\
    \text{ and } & \alpha_i = 1 - \exp(-\sigma_i(s_{i+1}-s_i)).
\end{align}
Here, $\{s_i\}_{i=1}^N$ is a set of samples from near bound $s_n$ to far bound $s_f$ and $\density_i=\density(\mathbf{r}(s_i))$, $\radiance_i=\radiance(\mathbf{r}(s_i))$ are evaluations of volume density and radiance at sample points $\mathbf{r}(s_i)$ along the ray.

Equation \eqref{eq:static_radiance} can only represent time-invariant scenes. 
Similar to methods using a MLP to model time-dependent functions \cite{sitzmann:etal:arXiv20}, a straightforward extension to represent dynamic scenes using NeRF is to concatenate time $t \in \R$ to the input of the implicit function $\mlp$ as
\begin{align}
    \mlp^{t} : \mathbf{r}(s), \mathbf{d}, t \rightarrow \density, \radiance
\end{align}
However, this time-dependent extension neither accurately reconstructs the complex time-varying scene using real-world images nor provides dynamic scene factorization to support animation, as we will show this in \secref{sec:experiments}.

\paragraph{Rigidly composed dynamic radiance fields}
Instead, we represent the scene using two time-invariant implicit volumetric models: a static NeRF $\mlp^{S}$ for static components and a dynamic NeRF $\mlp^{D}$ for dynamic components.
\begin{align}
    F_\theta^{S} &: \mathbf{r}(s_i), \mathbf{d} \rightarrow \density^{S}_{i}, \radiance^{S}_{i} \\
    F_\theta^{D} &: \mathbf{r}(s_i), \mathbf{d}, \xi(t) \rightarrow \density^{D}_{i}, \radiance^{D}_{i}
\end{align}
Here, $\xi(t) \in \se{3}$ is a set of time-dependent rigid poses that defines the transformation $T_{ow}(t)=\exp(\xi(t)) \in \SE{3}$ in world coordinate from time $t$ to time 0, which aligns the dynamic volume and the static volume under a single time-invariant canonical frame.

To compute the color of a pixel 
at a particular time $t$, we compose the transformed dynamic NeRF with the static NeRF using $\xi(t)$ and alpha blending. Specifically, given a set of samples $\{s_i\}_{i=1}^N$, we evaluate the static density and radiance $\sigma^S_i,\mathbf{c}^S_i$ at the points $\mathbf{r}(s_i)$, as well as the dynamic density and radiance $\sigma^D_i,\mathbf{c}^D_i$ at the \emph{transformed} points $T_{ow}(t)\mathbf{r}(s_i)$. We derive the compositional radiance $\hat{C}(\ray)$ based on \eqref{eq:static_radiance} as
\begin{align}
    & \hat{C}(\mathbf{r}) = \sum_{i=1}^NT_i(\alpha^S_i\mathbf{c}^S_i+\alpha^D_i\mathbf{c}^D_i) \label{eq:composed_radiance} \\
    \text{where } & T_i = \exp\left(-\sum_{j=1}^{i-1}(\sigma^S_j+\sigma^D_j)(s_{j+1}-s_j)\right) \\
    \text{ and } & \alpha^S_i = 1 - \exp(-\sigma^S_i(s_{i+1}-s_i)).
\end{align}
%
%

Our rigidly composed NeRF representation decouples rigid motion from geometry and appearance, allowing it to have full control over the dynamics of the environment. As shown in Sec.~\ref{sec:anim}, this enables applications such as removing the dynamic object and animating novel trajectories which cannot be achieved via a simple time-varying model.

\paragraph{Architecture}
We use the same MLP $\mlp$ model as \cite{mildenhall:etal:arXiv20} for both static and dynamic NeRF.
We also use positional encoding with the same bandwidth for the inputs and hierarchical volume sampling strategy. We use two independent coarse and fine MLP models to represent static and dynamic NeRFs. The coarse models use stratified sampling along the ray while the fine model uses importance sampling where the importance weights are the composed density $\sigma^S_j+\sigma^D_j$ obtained by both coarse models. 
This ensures the same sample intervals $\{s_i\}_{i=1}^N$ for the static and dynamic NeRFs, which enables the field composition shown in \eqref{eq:composed_radiance}. 

While we choose NeRF as the underlying 3D representation due to its simplicity and excellent performance, it is worth noting that STaR as a differentiable renderer can generally subsume any 3D static scene representation as $\mlp$ which shares the same input and output mapping.

\subsection{Optimizing STaR}
\label{sec:optimization}

During training, we optimize the following objectives
\begin{align}
    \mathcal{L} &= \sum_{\mathbf{r}\in\mathcal{R}}\left(\|\hat{C}_c(\mathbf{r})-C(\mathbf{r})\|^2-\|\hat{C}_f(\mathbf{r})-C(\mathbf{r})\|^2\right) \label{eq:mse} \\
    &+ \beta\sum_{i=1}^M\left(\mathcal{H}(\alpha^S_i) + \mathcal{H}(\alpha^D_i) + \mathcal{H}(\alpha^S_i, \alpha^D_i)\right) \label{eq:entropy_regularization}
\end{align}
The first term in \eqref{eq:mse} is a MSE loss, where $\hat{C}_c(\mathbf{r})$ and $\hat{C}_f(\mathbf{r})$ are the radiance (RGB) rendered by the coarse and fine model respectively, $C(\mathbf{r})$ is the ground truth color and $\mathcal{R}$ is the set of rays in a batch. Note that $C(\mathbf{r})$ is the only source of supervision.
Given a limited number of camera views, minimizing the objective in \eqref{eq:mse} can suffer from local optima. 
We introduce the regularization term in \eqref{eq:entropy_regularization} which we observe can converge better.

\paragraph{Entropy regularization}
\eqref{eq:entropy_regularization} regularizes the entropy $\mathcal{H}$ of the rendered transparency values and is summed over all samples along all rays in a batch. It consists of two parts. The first part $\mathcal{H}(\alpha^S_i) + \mathcal{H}(\alpha^D_i)$ encourages the transparencies to be close to 0 or 1, where
\begin{equation}
    \mathcal{H}(\alpha^S_i) = \alpha^S_i\log{\alpha^S_i} + (1-\alpha^S_i)\log(1-\alpha^S_i)
\end{equation}
and $\mathcal{H}(\alpha^D_i)$ is computed likewise. This helps reducing fuzziness in the volume.
The second part $\mathcal{H}(\alpha^S_i, \alpha^D_i)$ prevents the static and dynamic volume from both having large density at the same point, which helps the model to obtain a less entangled decomposition. Specifically, 
\begin{equation}
    \mathcal{H}(\alpha^S_i, \alpha^D_i) = \left(\overline{\alpha}^S_i\log{\overline{\alpha}^S_i} + \overline{\alpha}^D_i\log{\overline{\alpha}^D_i}\right)(\alpha^S_i+\alpha^D_i)
\end{equation}
where $\overline{\alpha}^S_i=\alpha^S_i/(\alpha^S_i+\alpha^D_i)$ is the normalized transparency. The term is weighted by the total transparency $\alpha^S_i+\alpha^D_i$, so that a point is allowed to be empty but not occupied in both static and dynamic volume.



\paragraph{Rigid pose optimization}
We define the canonical frame by assigning the pose $\xi(0)$ at time 0 to the identity. We optimize pose parameters over time $\xi(t) \in \se{3}$ on the Lie algebra \cite{gabay1982minimizing}.
During optimization, we represent the transformation from dynamic to static volume at time $t$ as an iterative update $\exp(\epsilon + \xi(t))$, where $\epsilon \in \se{3}$ is a local perturbation on the manifold that is initialized to zero at the beginning of each iteration. We can compute the gradient of $\epsilon$ using the gradient of the transformed point $\exp(\epsilon + \xi(t))\mathbf{p}$ and an analytical Jacobian that can be computed from $\xi(t)$ in the forward pass. Then, we update $\xi(t)$ as $\xi(t)=a_\xi\nabla\epsilon + \xi(t)$.
Note that we use different learning rate $a_\xi$ for the pose parameters and $a_\theta$ for the NeRF parameters. We include the full analytic form of Jacobian in the supplementary material.


\paragraph{Appearance initialization}
We observe there are many local optima in the optimization of STaR due to entanglement of geometry, appearance and pose, which requires careful initialization. To initialize, we first train a static NeRF $\mlp^{S}$ following \eqref{eq:static_radiance} only using the images from the first frame.
The initialization is terminated when the average MSE of the fine model $\tfrac{1}{|\mathcal{R}_1|}\sum_{r\in\mathcal{R}_1}\|\hat{C}_f(\mathbf{r})-C(\mathbf{r})\|$ over all images from the first frame $\mathcal{R}_1$ is below the threshold $m_1$. 
Note that at this stage, the static volume can (and likely does) contain information from the dynamic object, but we observe that this initialization provides the model with a good initial estimate of the geometry and appearance from which to begin disentangling the scene.

\paragraph{Online training}
We train STaR in an online fashion and can handle video sequences of arbitrary length. 
After appearance initialization, the static NeRF $\mlp^{S}$, dynamic NeRF $\mlp^{D}$ and pose parameters $\{\xi_t\}_{t=1}^k$ are optimized jointly on the first $k$ frames of the video. $k$ is set to $k_0$ initially and incremented when the average MSE of the fine model $\tfrac{1}{|\mathcal{R}_k|}\sum_{r\in\mathcal{R}_{k}}\|\hat{C}_f(\mathbf{r})-C(\mathbf{r})\|$ over all rays in the first $k$ frames $\mathcal{R}_{k}$ is below the threshold $m_2$. For $t\leq k_0$, $\xi(t)$ is initialized to identity and for $t>k_0$, $\xi(k)$ is set to the pose of the previous frame $\xi(k-1)$ when frame $k$ is added. 
In \secref{sec:experiments}, we show that both appearance initialization and online training is crucial for the optimization of STaR.

\section{Experiments}
\label{sec:experiments}

Our experiments demonstrate that STaR is able to decouple static and dynamic components of a dynamic scene in challenging synthetic as well as real-world scenarios (see \figref{fig:novel_view_comparison}). 
As a result, STaR is not only capable of synthesizing a dynamic scene from novel views and time with superior quality (see \secref{sec:nvs}) but also animating an imaginary trajectory in a photorealistic fashion (see \secref{sec:anim}). 

We implement STaR in PyTorch. We summarize the set of hyperparameters and training time/GPU used in \tabref{tab:hyperparams}, which includes the batch size $B$, the entropy regularization weight $\beta$, learning rates $a_\theta,a_\xi$ for NeRF and pose parameters, MSE thresholds $m_1,m_2$ for appearance initialization and online training, and number of bootstrapping frames $k_0$. We optimize the objective using ADAM. Following \cite{mildenhall:etal:arXiv20}, we exponentially decay $a_\theta$ every 250K steps with decay rate $\gamma$, but we do not decay $a_\xi$. 

\begin{table}[t]
    \centering
    \scriptsize
    \tabcolsep 2.5pt
    \begin{tabular}{c|cccccccccc}
        \toprule
        Sequence & $B$ & $\beta$ & $a_\theta$ & $\gamma$ & $a_\xi$ & $m_1$ & $m_2$ & $k_0$ & GPU & Time \\ \midrule
        Lamp and desk & 3200 & 2e-3 & 5e-4 & 0.5 & 5e-5 & 4e-4 & 2e-4 & 5 & 2 & 12hr \\
        Kitchen table & 3200 & 2e-3 & 5e-4 & 0.5 & 5e-5 & 5e-4 & 5e-4 & 5 & 2 & 24hr \\
        Moving banana & 10816 & 2e-3 & 5e-4 & 0.8 & 5e-4 & 2e-3 & 1e-3 & 5 & 8 & 55hr \\
        \bottomrule
    \end{tabular}
    \caption{\small Hyperparameters and training time.}
    \label{tab:hyperparams}
    \vspace{-1em}
\end{table}

\begin{table*}[h!]
    \centering
    \footnotesize
    \begin{tabular}{cc}
        \rotatebox[origin=c]{90}{Composition} & 
        \begin{tabular}{cccccccccc}
            \toprule
            Sequence & \multicolumn{3}{c}{Lamp and desk (synthetic)} & \multicolumn{3}{c}{Kitchen table (synthetic)} & \multicolumn{3}{c}{Moving banana (real)} \\ \cmidrule(lr){2-4} \cmidrule(lr){5-7} \cmidrule{8-10}
            Metric & PSNR $\uparrow$ & SSIM $\uparrow$ & LPIPS $\downarrow$ & PSNR $\uparrow$ & SSIM $\uparrow$ & LPIPS $\downarrow$ & PSNR $\uparrow$ & SSIM $\uparrow$ & LPIPS $\downarrow$ \\ \midrule
            NeRF \cite{mildenhall:etal:arXiv20} & 22.09 & 0.873 & 0.182 & 20.39 & 0.657 & 0.374 & 24.48 & 0.777 & 0.261 \\
            NeRF-time & 25.86 & 0.873 & 0.101 & 21.95 & 0.686 & 0.319 & 24.82 & 0.776 & 0.259 \\
            NeRF-W \cite{martin:etal:arXiv20} & 27.68 & 0.931 & 0.048 & 27.99 & 0.749 & 0.220 & \textbf{27.26} & 0.791 & 0.257 \\
            STaR (ours) & \textbf{32.95} & \textbf{0.957} & \textbf{0.023} & \textbf{29.51} & \textbf{0.767} & \textbf{0.195} & 27.19 & \textbf{0.803} & \textbf{0.209} \\
            \bottomrule
        \end{tabular} \\
        \rotatebox[origin=c]{90}{Static} & 
        \begin{tabular}{cccccccccc}
            \toprule
            Sequence & \multicolumn{3}{c}{Lamp and desk (synthetic)} & \multicolumn{3}{c}{Kitchen table (synthetic)} & \multicolumn{3}{c}{Moving banana (real)} \\ \cmidrule(lr){2-4} \cmidrule(lr){5-7} \cmidrule{8-10}
            Metric & PSNR $\uparrow$ & SSIM $\uparrow$ & LPIPS $\downarrow$ & PSNR $\uparrow$ & SSIM $\uparrow$ & LPIPS $\downarrow$ & PSNR $\uparrow$ & SSIM $\uparrow$ & LPIPS $\downarrow$ \\ \midrule
            NeRF \cite{mildenhall:etal:arXiv20} & 25.82 & 0.921 & 0.092 & 21.02 & 0.667 & 0.353 & 25.54 & 0.793 & 0.235 \\
            NeRF-time & 30.00 & 0.925 & 0.064 & 22.56 & 0.695 & 0.309 & 25.51 & 0.790 & 0.244 \\
            NeRF-W \cite{martin:etal:arXiv20} & 34.12 & 0.971 & 0.025 & 29.25 & 0.760 & 0.203 & \textbf{27.68} & 0.801 & 0.243 \\
            STaR (ours) & \textbf{34.57} & \textbf{0.967} & \textbf{0.018} & \textbf{29.75} & \textbf{0.771} & \textbf{0.193} & 27.44 & \textbf{0.812} & \textbf{0.198} \\
            \bottomrule
        \end{tabular} \\
        \rotatebox[origin=c]{90}{Dynamic} & 
        \begin{tabular}{cccccccccc}
            \toprule
            Sequence & \multicolumn{3}{c}{Lamp and desk (synthetic)} & \multicolumn{3}{c}{Kitchen table (synthetic)} & \multicolumn{3}{c}{Moving banana (real)} \\ \cmidrule(lr){2-4} \cmidrule(lr){5-7} \cmidrule{8-10}
            Metric & PSNR $\uparrow$ & SSIM $\uparrow$ & LPIPS $\downarrow$ & PSNR $\uparrow$ & SSIM $\uparrow$ & LPIPS $\downarrow$ & PSNR $\uparrow$ & SSIM $\uparrow$ & LPIPS $\downarrow$ \\ \midrule
            NeRF \cite{mildenhall:etal:arXiv20} & 14.33 & 0.498 & 0.464 & 10.83 & 0.339 & 0.319 & 16.87 & 0.336 & 0.567 \\
            NeRF-time & 18.48 & 0.555 & 0.212 & 13.80 & 0.469 & 0.243 & 20.68 & 0.477 & 0.341 \\
            NeRF-W \cite{martin:etal:arXiv20} & 19.04 & 0.597 & 0.142 & 15.54 & 0.490 & 0.230 & 24.71 & \textbf{0.687} & 0.298 \\
            STaR (ours) & \textbf{28.11} & \textbf{0.903} & \textbf{0.027} & \textbf{23.97} & \textbf{0.837} & \textbf{0.036} & \textbf{25.14} & \textbf{0.687} & \textbf{0.228} \\
            \bottomrule
        \end{tabular}
    \end{tabular}
    \caption{\small \textbf{Quantitative comparison of our method to baselines in photorealistic interpolation of motion.} The models are evaluated on a 4x slow motion of the original video from a fixed novel viewpoint. The synthetic data are followed with a notation (synthetic), and the real world data is notated as (real). We use a ground truth bounding box as a reference to evaluate static and dynamic regions separately.}
    \label{tab:quantitative_all}
    \vspace{-1em}
\end{table*}

\paragraph{Datasets} 
We created two synthetic and one real world multi-view RGB videos to evaluate STaR. The videos are highly challenging, containing complex geometry, large object motion, significant view-dependent and time-dependent visual effects such as specular highlight and shadows, all of which are not present in any existing public dataset. Please refer to supplementary materials for more details.
\begin{itemize} [noitemsep, leftmargin=*]
    \item \textbf{Synthetic data:} We rendered two synthetic videos using Blender: \emph{lamp and desk}, a 15-frame video showing a chair pulled away from the desk in a study room, and \emph{kitchen table}, a 20-frame video showing a vase sliding across a reflective breakfast bar in a kitchen. The motion is created by modifying the pose of an object in a photorealistic scene. We leave the material and lighting produced by human designers untouched. Both videos have 8 fixed camera views for training and 1 held-out view for evaluation with an image resolution of $400\times400$. 
    \item \textbf{Real world data:} We captured a multi-view video, \emph{moving banana}, in real natural indoor environment. In the scene, a moving banana toy is put on a robot vacuum which sweeps across a living room. The data consists of 17 time-synchronized videos with a total of 792 temporal frames. To demonstrate our algorithm on relatively large motion, we uniformly subsample the video to 38 key frames and use 16 views for training and 1 held-out view for evaluation. The image resolution is $676\times507$.
\end{itemize}

\subsection{4D Novel View Synthesis Evaluation}
\label{sec:nvs}

We first evaluate our approach on its ability to photorealisticly render the dynamic scene from a novel 4D view (3D camera pose + time). More specifically, we render a 4x slow motion of the original video from a novel camera view by linearly interpolating the estimated object poses. We compare our method to the following baselines (see supplementary for the detailed architecture):
\vspace{-0.5em}
\begin{itemize} [noitemsep, leftmargin=*]
\item \textbf{NeRF \cite{mildenhall:etal:arXiv20}:} The original NeRF assumes the scene is static. This baseline cannot properly reconstruct the dynamic object, but can still provide a reference to the quality in terms of the static part of the scenes.
\item \textbf{NeRF-time:} This model takes positional-encoded time as an additional input, thus creating a 4D representation using a strategy similar to \cite{sitzmann:etal:arXiv20, tancik:etal:neuralips20}.
\item \textbf{NeRF-W \cite{martin:etal:arXiv20}:} This model takes an additional latent code input as proposed by Martin-Brualla \etal \cite{martin:etal:arXiv20}. We did not include the appearance embedding as did NeRF-W in \cite{martin:etal:arXiv20} because our video data does not contain strong appearance variations. This is the most related existing work that can handle dynamic variation based on NeRF.
\end{itemize}
\vspace{-0.5em}

\begin{figure}[ht]
    \centering
    \tabcolsep 3.5 pt
    \begin{tabular}{ccc}
        \includegraphics[width=0.25\linewidth]{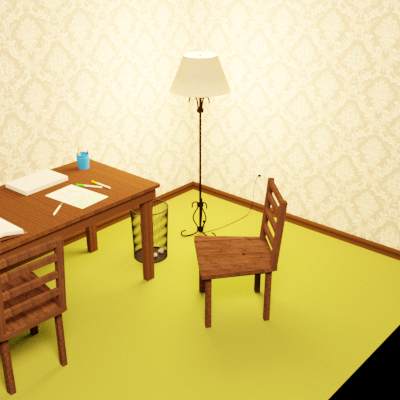} &
        \includegraphics[width=0.25\linewidth]{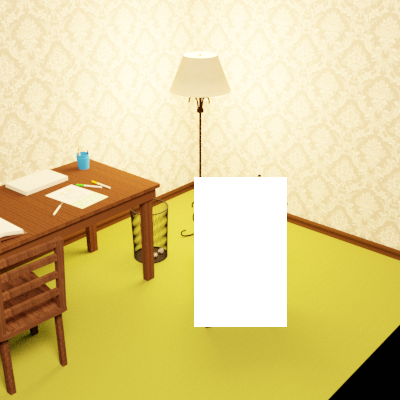} &
        \includegraphics[width=0.15\linewidth]{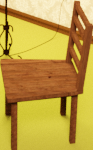} \\
    \end{tabular}
    \caption{\small A visual illustration of our evaluation protocol (from left to right: Composition, Static, Dynamic).}
    \label{fig:evaluation}
    \vspace{-1em}
\end{figure}

We evaluate the novel view synthesis quality in terms of photorealism following the standard metrics in \cite{mildenhall:etal:arXiv20}: the average PSNR, SSIM and LPIPS \cite{zhang:etal:cvpr2018} across all test frames. Since our baselines do not provide separate static and dynamic rendering, we use object bounding boxes to divide the image into static background and dynamic foreground (see \figref{fig:evaluation}). We compute metrics on the original image, the static background and the dynamic foreground.

\begin{figure*}[h!]
    \centering
    \tabcolsep 1.5pt
    \begin{tabular}{ccc|cc}
        & Key frame & Interpolated frame & Key frame & Interpolated frame \\
        \rotatebox[origin=c]{90}{NeRF-time} & 
        \raisebox{-.5\height}{\includegraphics[width=0.23\linewidth]{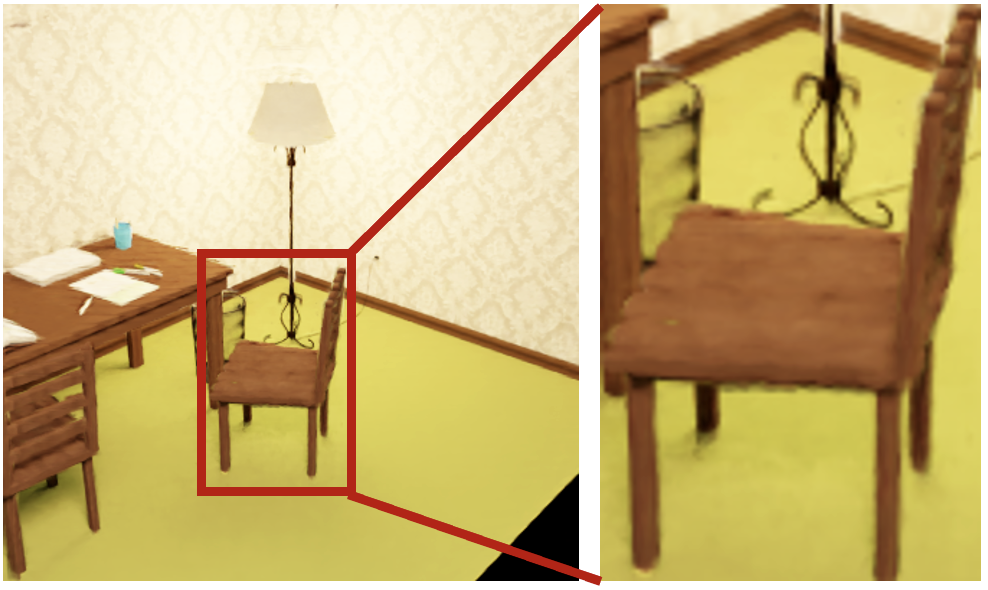}} & 
        \raisebox{-.5\height}{\includegraphics[width=0.23\linewidth]{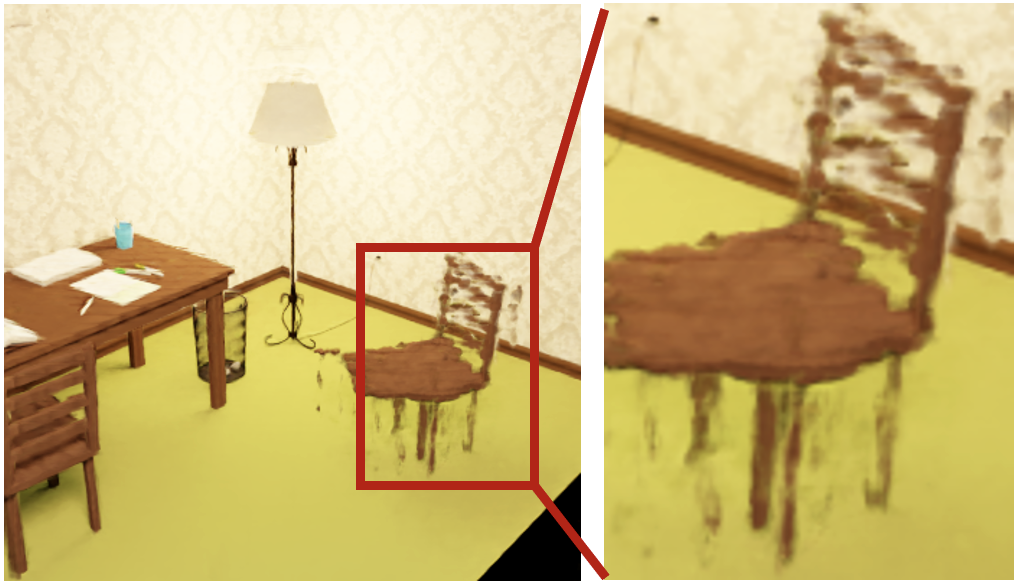}} & 
        \raisebox{-.5\height}{\includegraphics[width=0.23\linewidth]{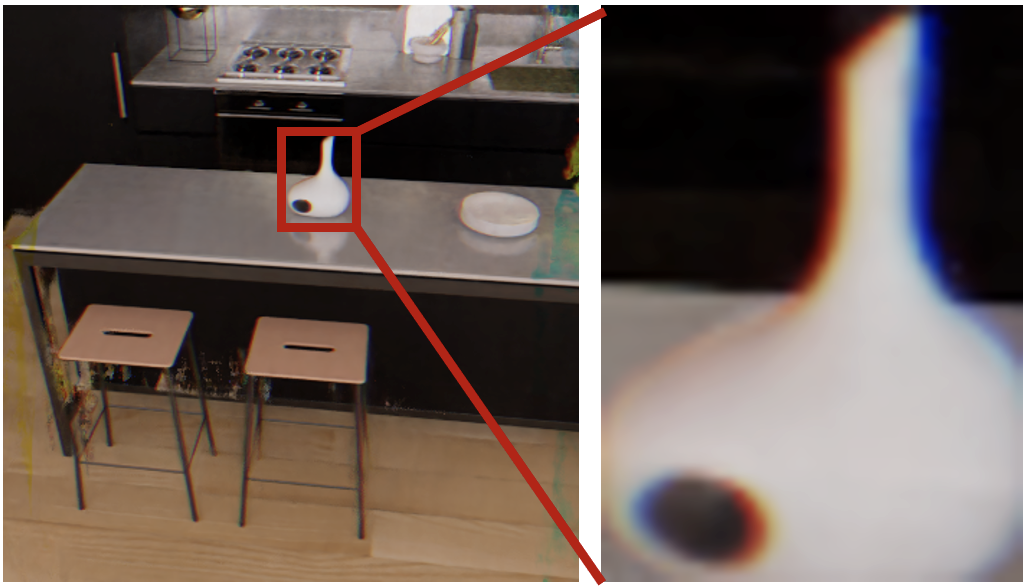}} & 
        \raisebox{-.5\height}{\includegraphics[width=0.23\linewidth]{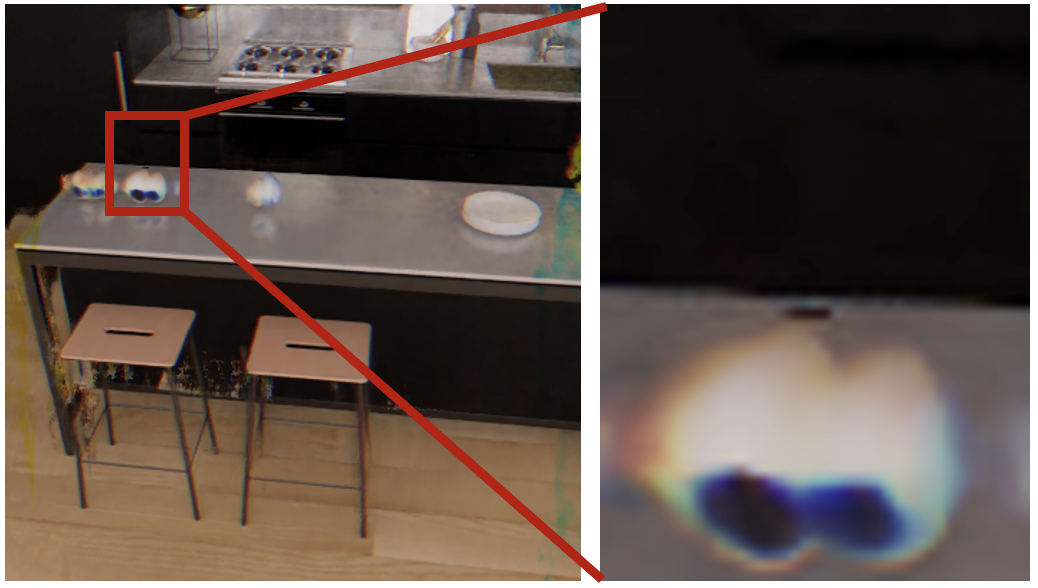}} \\
        \rotatebox[origin=c]{90}{NeRF-W\cite{martin:etal:arXiv20}} & 
        \raisebox{-.5\height}{\includegraphics[width=0.23\linewidth]{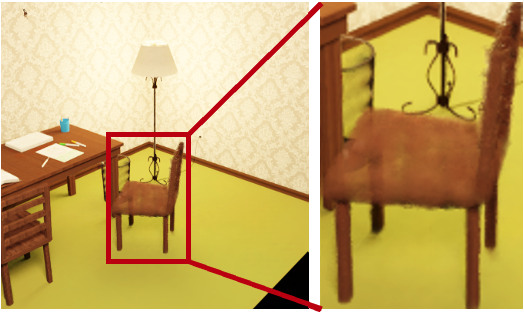}} & 
        \raisebox{-.5\height}{\includegraphics[width=0.23\linewidth]{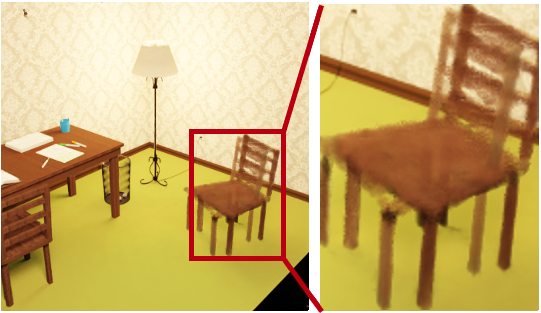}} & 
        \raisebox{-.5\height}{\includegraphics[width=0.23\linewidth]{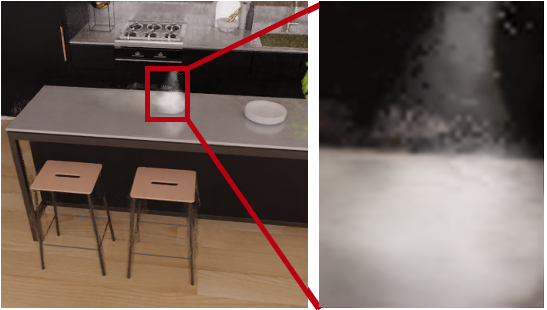}} & 
        \raisebox{-.5\height}{\includegraphics[width=0.23\linewidth]{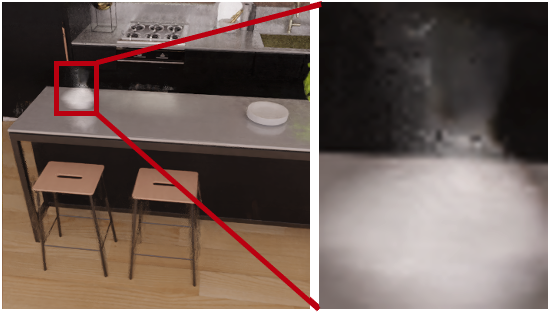}} \\
        \rotatebox[origin=c]{90}{STAR} & 
        \raisebox{-.5\height}{\includegraphics[width=0.23\linewidth]{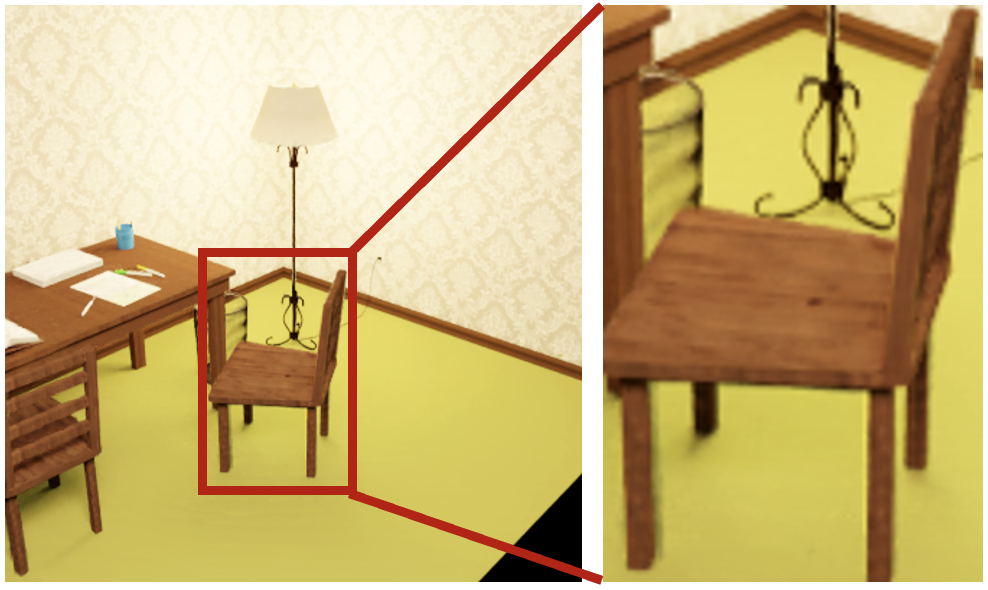}} & 
        \raisebox{-.5\height}{\includegraphics[width=0.23\linewidth]{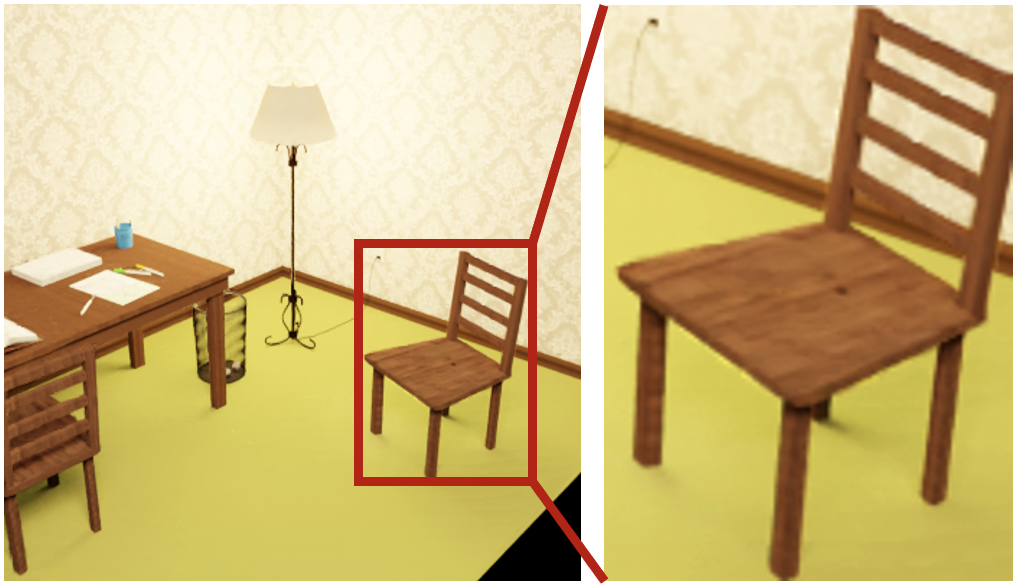}} & 
        \raisebox{-.5\height}{\includegraphics[width=0.23\linewidth]{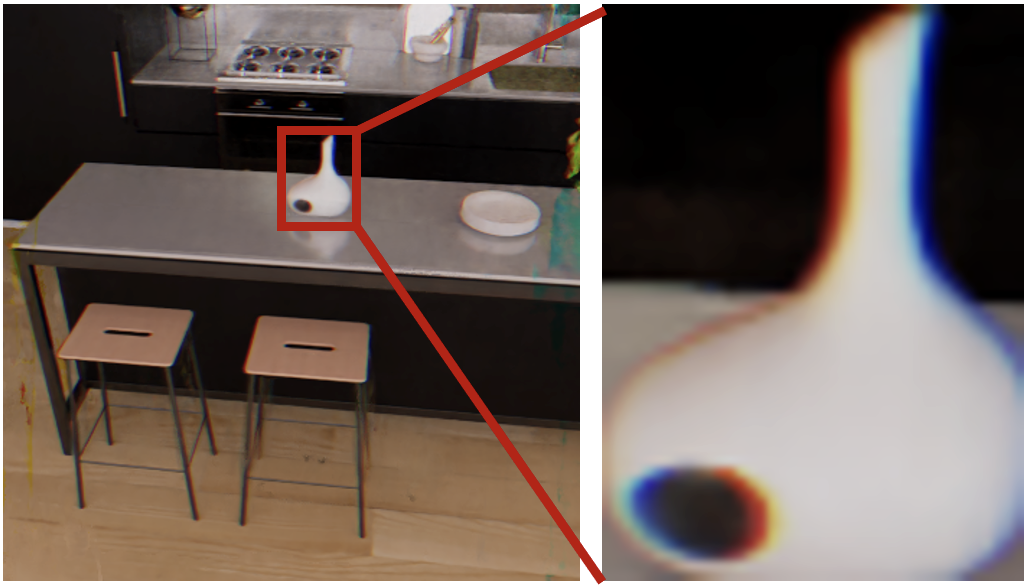}} & 
        \raisebox{-.5\height}{\includegraphics[width=0.23\linewidth]{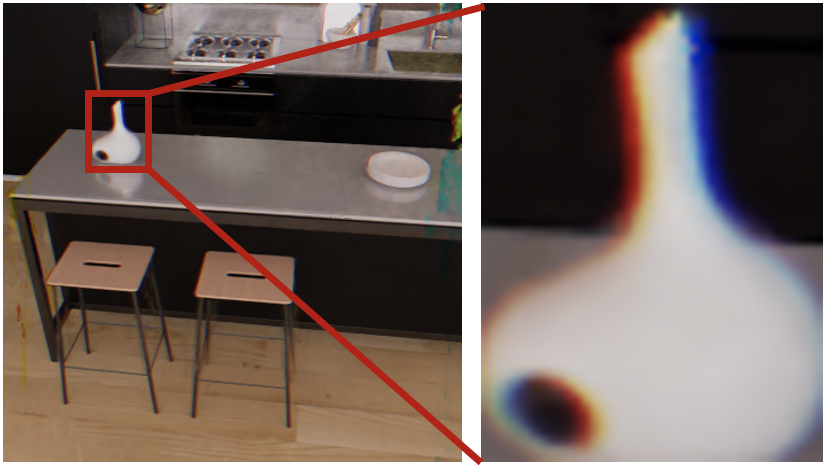}} \\
        \rotatebox[origin=c]{90}{Ground truth} & 
        \raisebox{-.5\height}{\includegraphics[width=0.23\linewidth]{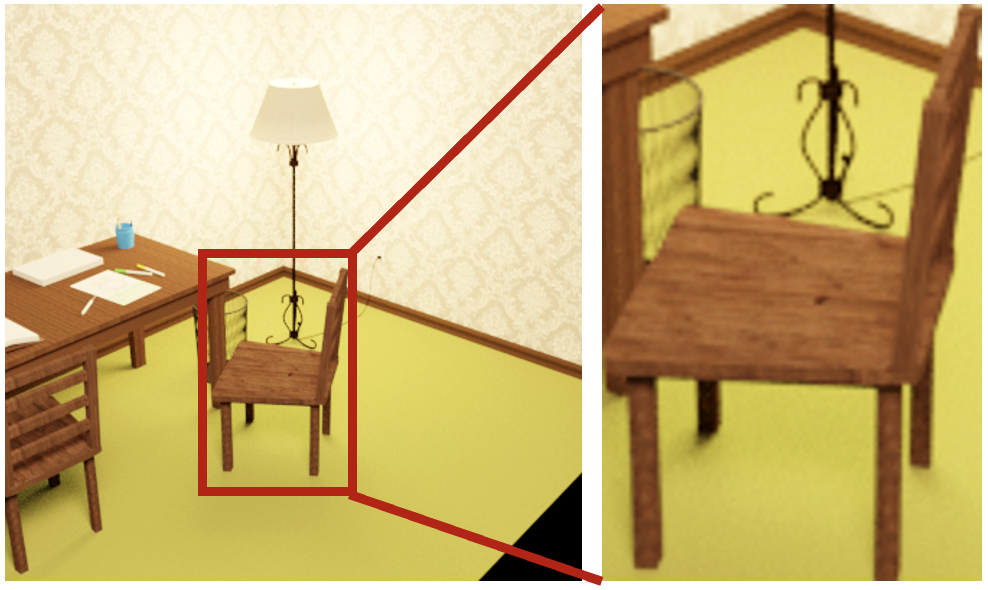}} & 
        \raisebox{-.5\height}{\includegraphics[width=0.23\linewidth]{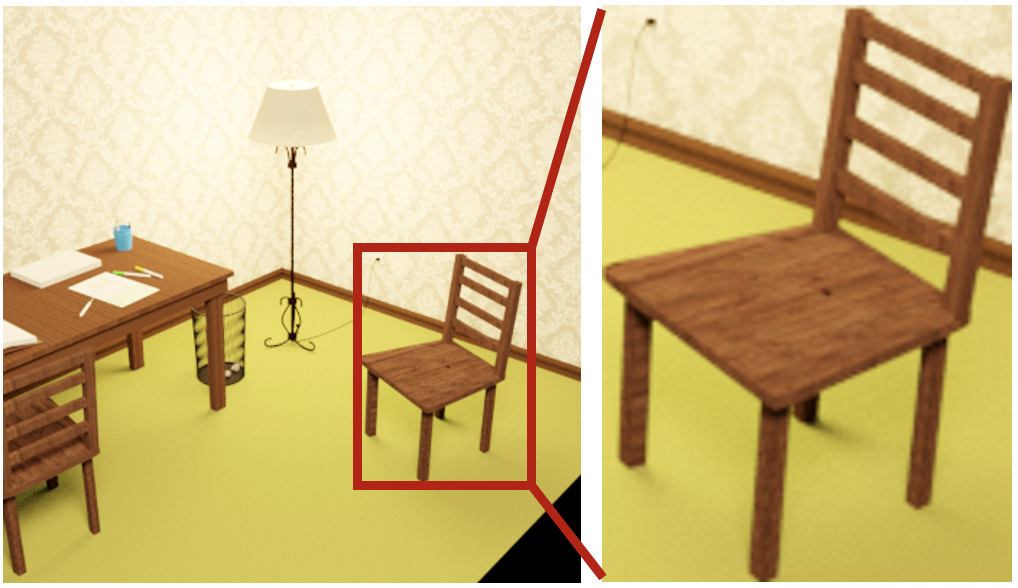}} & 
        \raisebox{-.5\height}{\includegraphics[width=0.23\linewidth]{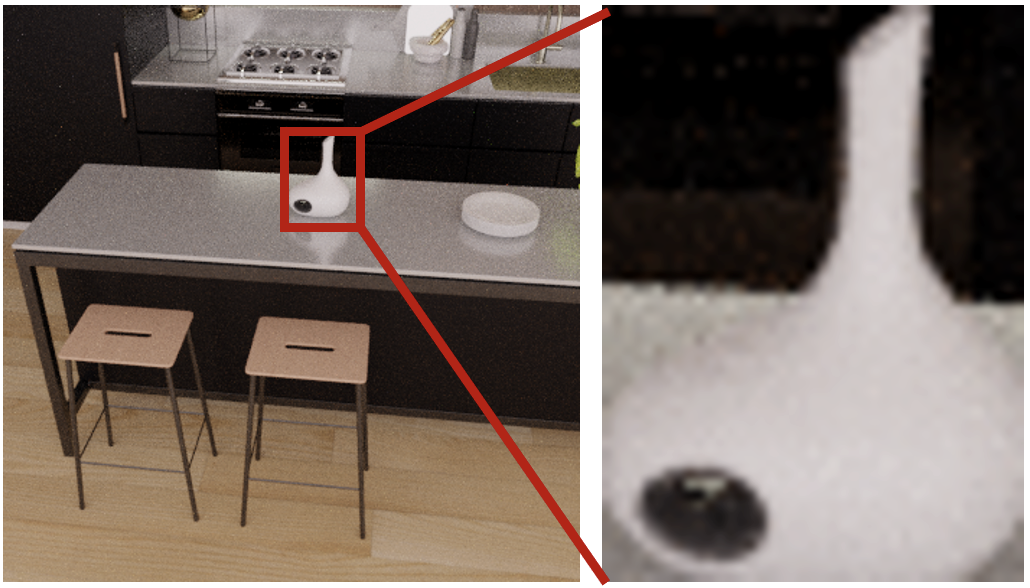}} & 
        \raisebox{-.5\height}{\includegraphics[width=0.23\linewidth]{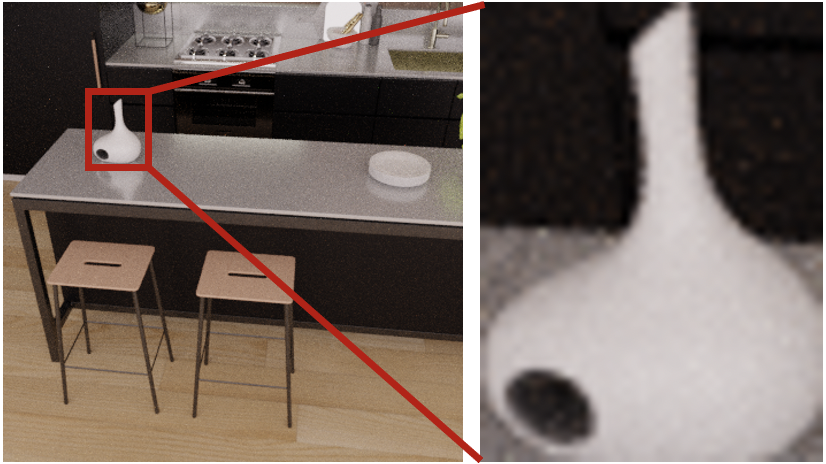}}
    \end{tabular}
    \begin{tabular}{cccc}
        \toprule
        & Key frame & \multicolumn{2}{c}{Interpolated frame} \\
         \rotatebox[origin=c]{90}{NeRF-time} & 
         \raisebox{-.5\height}{\includegraphics[width=0.3\linewidth]{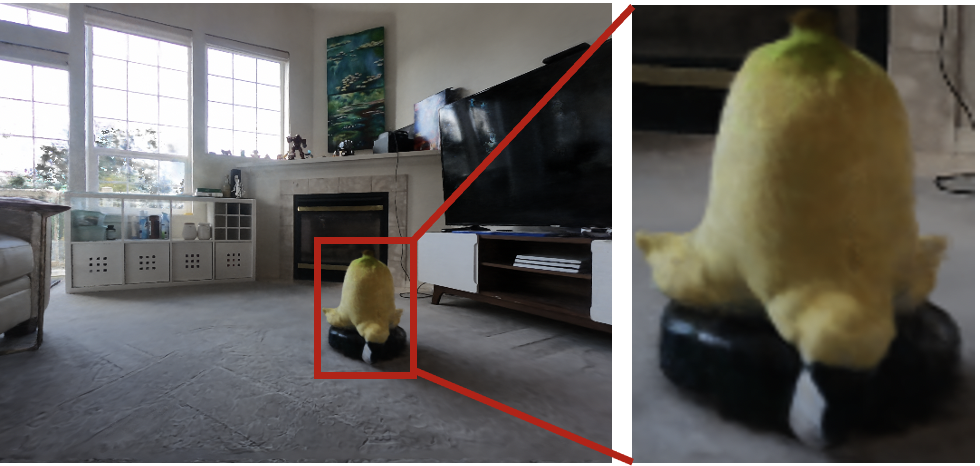}} & 
         \raisebox{-.5\height}{\includegraphics[width=0.32\linewidth]{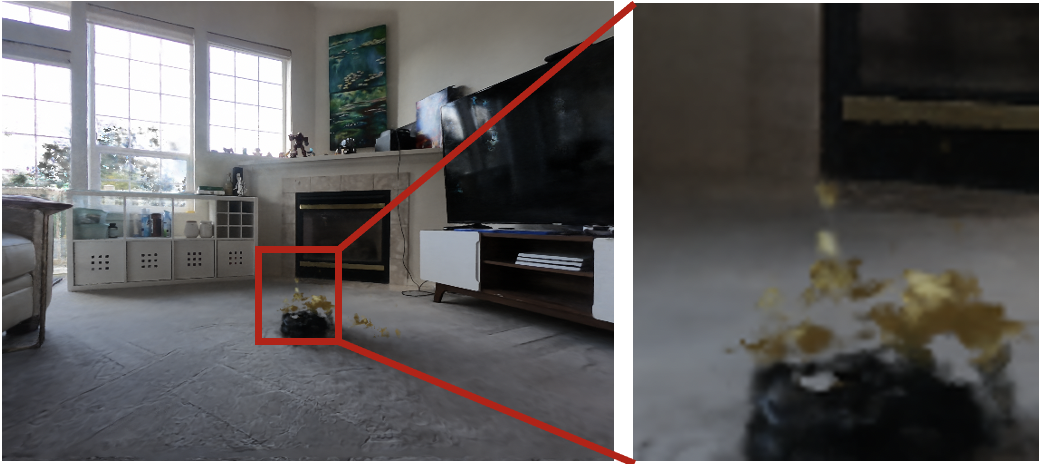}} &
         \raisebox{-.5\height}{\includegraphics[width=0.31\linewidth]{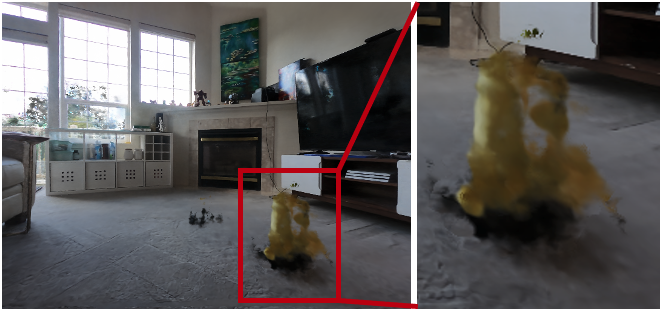}} \\
         \rotatebox[origin=c]{90}{NeRF-W \cite{martin:etal:arXiv20}} & 
         \raisebox{-.5\height}{\includegraphics[width=0.3\linewidth]{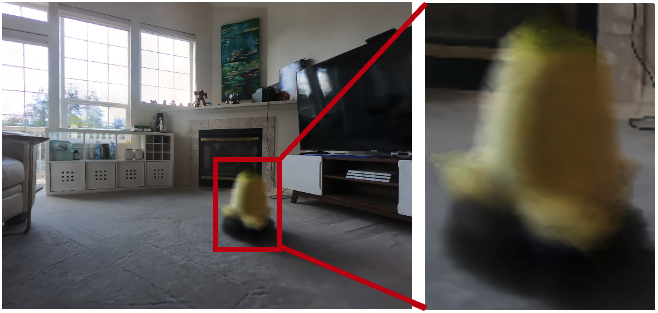}} & 
         \raisebox{-.5\height}{\includegraphics[width=0.32\linewidth]{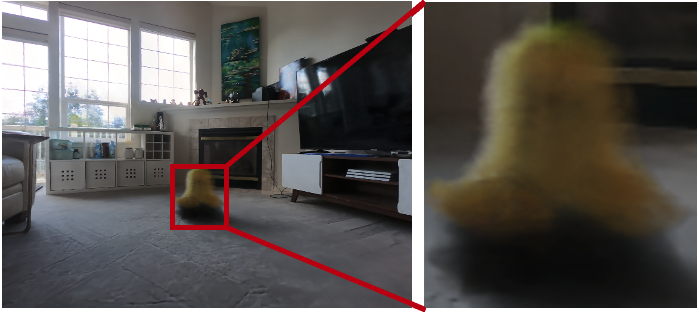}} &
         \raisebox{-.5\height}{\includegraphics[width=0.31\linewidth]{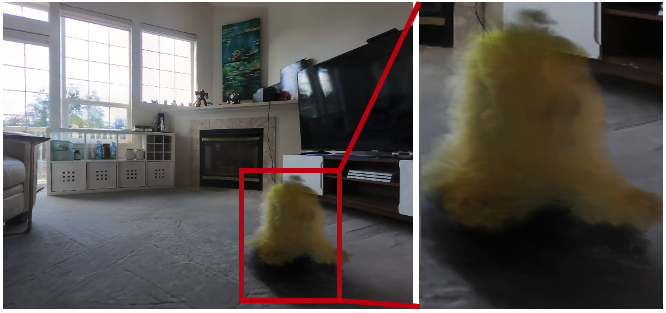}} \\
         \rotatebox[origin=c]{90}{STAR} & 
         \raisebox{-.5\height}{\includegraphics[width=0.3\linewidth]{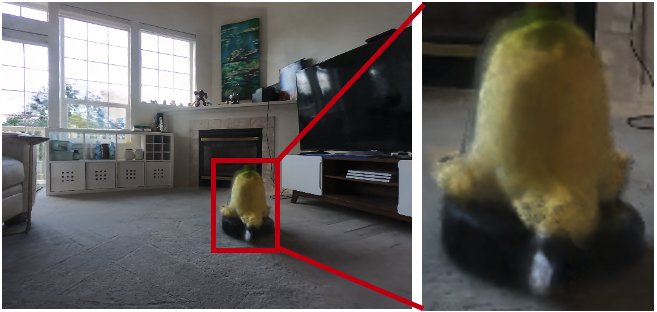}} & 
         \raisebox{-.5\height}{\includegraphics[width=0.32\linewidth]{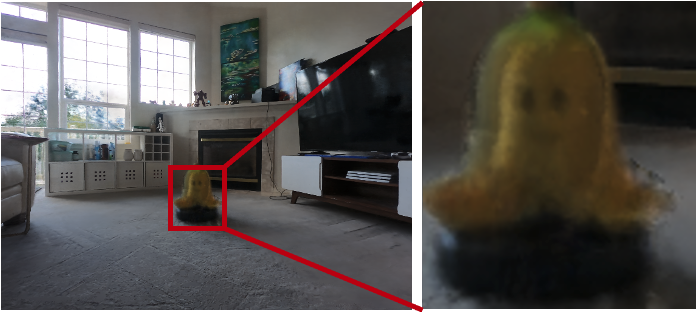}} &
         \raisebox{-.5\height}{\includegraphics[width=0.31\linewidth]{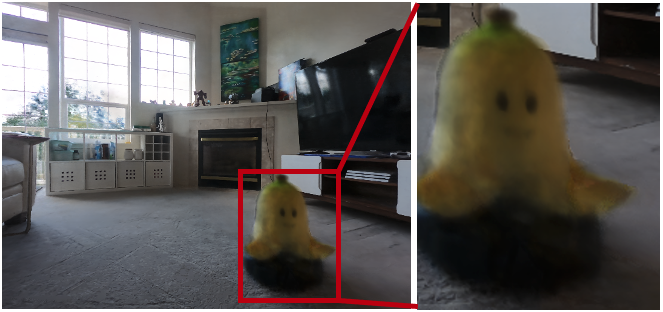}} \\
         \rotatebox[origin=c]{90}{Ground truth} & 
         \raisebox{-.5\height}{\includegraphics[width=0.3\linewidth]{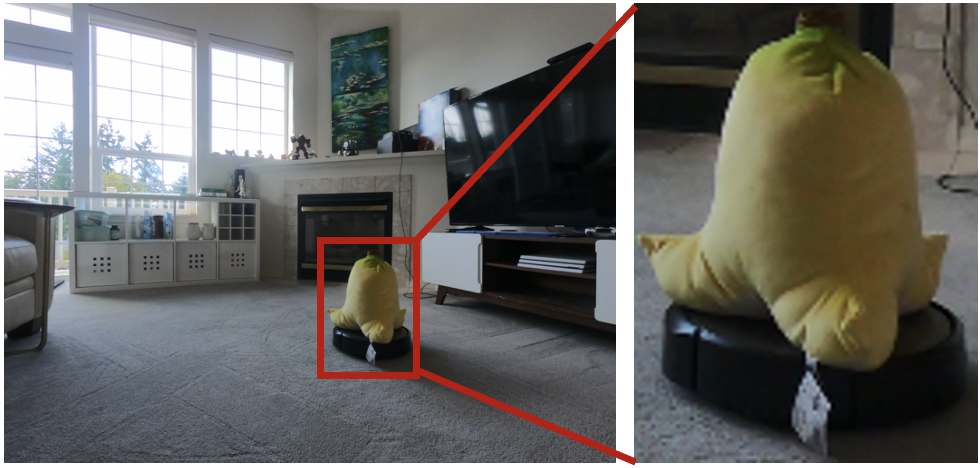}} & 
         \raisebox{-.5\height}{\includegraphics[width=0.32\linewidth]{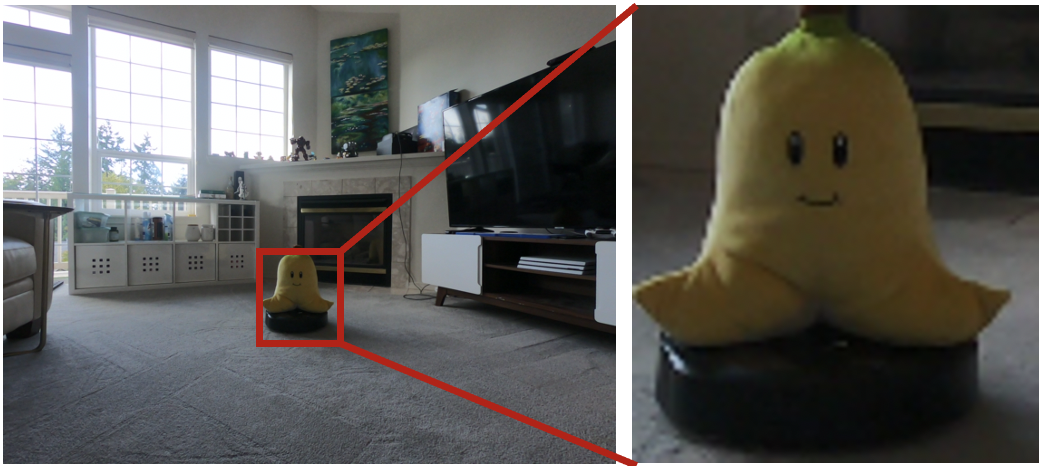}} &
         \raisebox{-.5\height}{\includegraphics[width=0.31\linewidth]{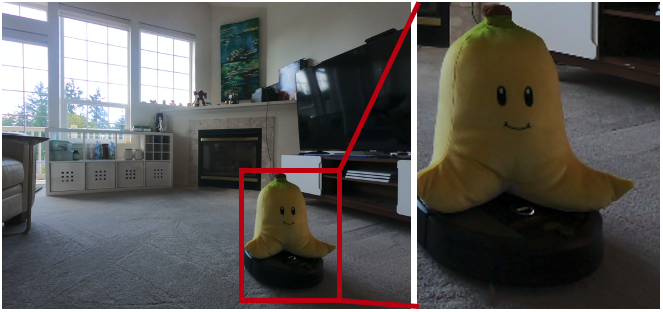}}
    \end{tabular}
    \caption{\small \textbf{Qualitative comparison of 4D novel view synthesis on both synthetic and real world data.} We compare synthesized novel views of the dynamic scene on key frames which share the same time stamps as training images, and on interpolated frames from time stamps that are not included in training (i.e. spatial-temporal novel views).}
    \label{fig:novel_view_comparison}
\end{figure*}
\begin{figure*}[h]
    \centering
    \tabcolsep 1.5pt
    \begin{tabular}{*5{c}}
         \raisebox{-.5\height}{\includegraphics[width=0.18\linewidth]{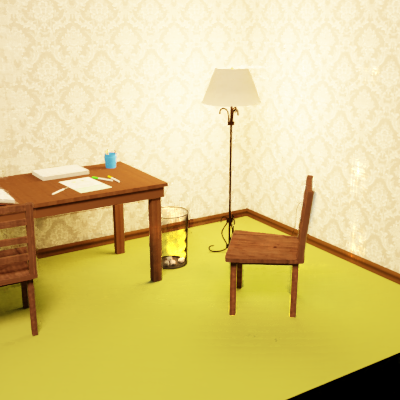}} & 
         \raisebox{-.5\height}{\includegraphics[width=0.18\linewidth]{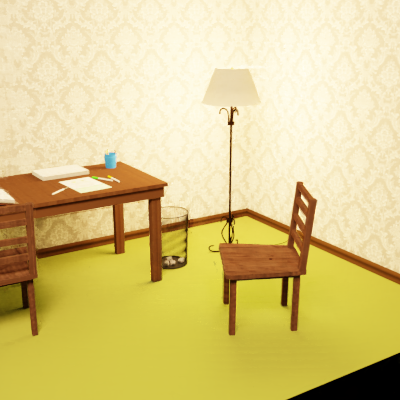}} & 
         \raisebox{-.5\height}{\includegraphics[width=0.18\linewidth]{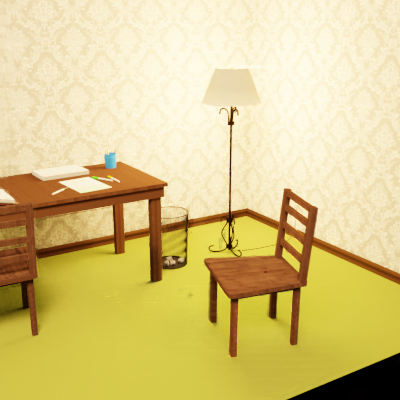}} & 
         \raisebox{-.5\height}{\includegraphics[width=0.18\linewidth]{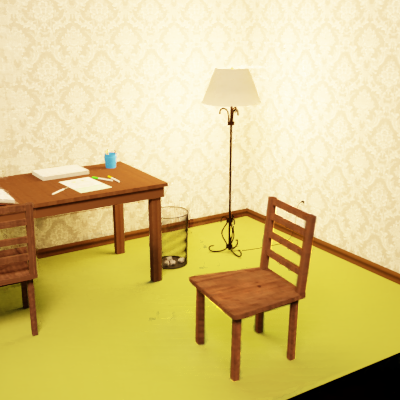}} & \raisebox{-.5\height}{\includegraphics[width=0.19\linewidth]{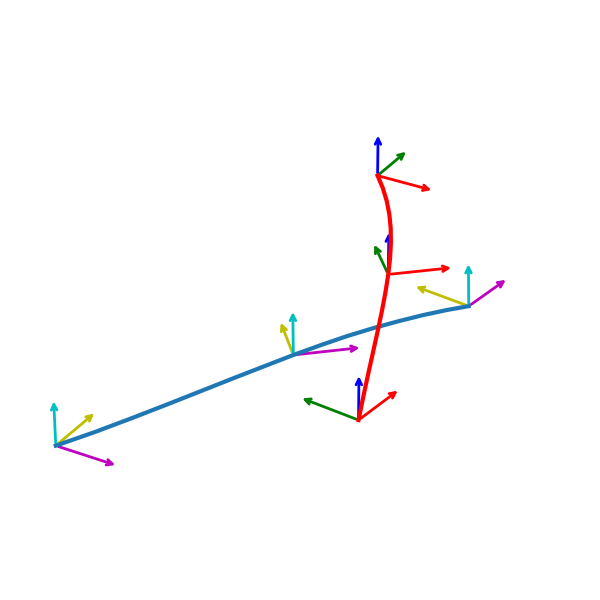}} \\
         \raisebox{-.5\height}{\includegraphics[width=0.18\linewidth]{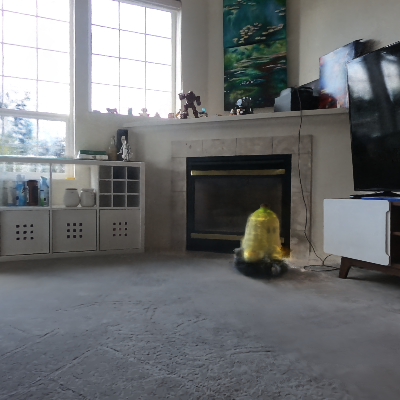}} & 
         \raisebox{-.5\height}{\includegraphics[width=0.18\linewidth]{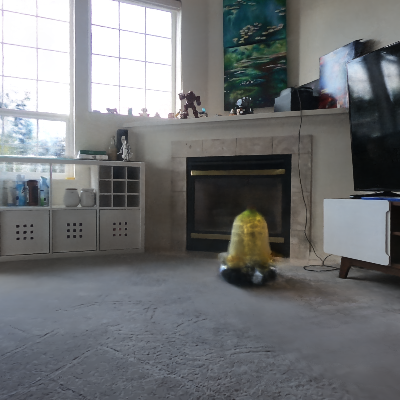}} & 
         \raisebox{-.5\height}{\includegraphics[width=0.18\linewidth]{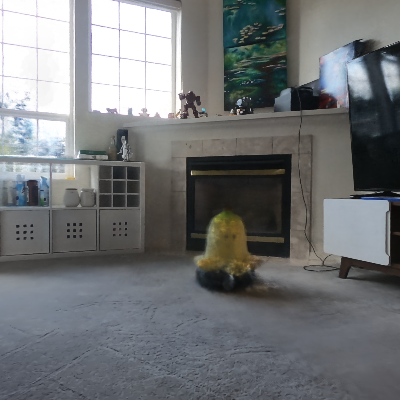}} & 
         \raisebox{-.5\height}{\includegraphics[width=0.18\linewidth]{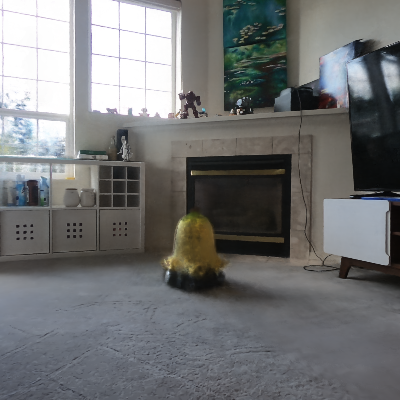}} & 
         \raisebox{-.5\height}{\includegraphics[width=0.19\linewidth]{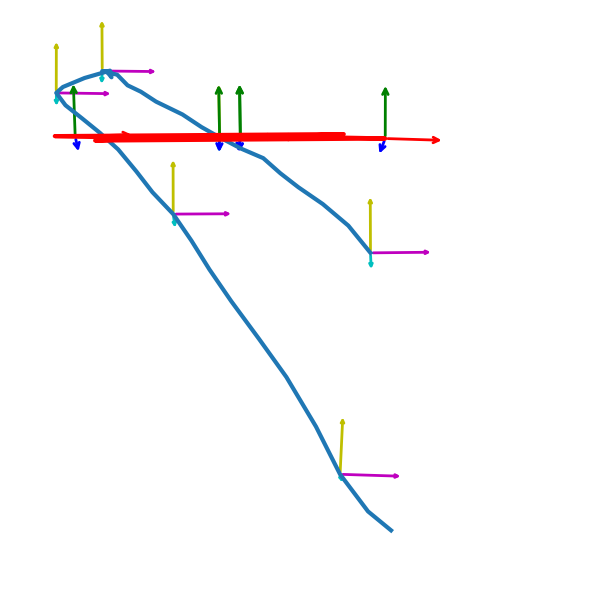}}
    \end{tabular}
    \caption{\small \textbf{Synthesizing novel trajectories}. We visualize synthesized novel views of the dynamic object moving along an imaginary trajectory unseen during training (rightmost column: \textcolor{blue}{training trajectory in blue} and \textcolor{red}{novel trajectory in red}) 
    }
    \label{fig:animation}
    \vspace{-1em}
\end{figure*}
\begin{table}[h!]
    \centering
    \footnotesize
    \begin{tabular}{lccccccccc}
        \toprule
         & PSNR $\uparrow$ & SSIM $\uparrow$ & LPIPS $\downarrow$  \\ \midrule
        No appr init* & 20.78 & 0.820 & 0.188 
        \\
        No online & 25.98 & 0.925 & 0.079 
        \\ 
        No entropy & 32.88 & \textbf{0.959} & 0.032 
        \\ \midrule
        Full model & \textbf{32.95} & 0.957 & \textbf{0.023} 
        \\
        \bottomrule
    \end{tabular}
    \caption{\small \textbf{Ablation study of optimization strategy discussed in \secref{sec:optimization} on the \emph{lamp and desk} sequence}. No appr init removes appearance initialization. No online indices training over entire video instead of online training. No entropy indicates only using MSE loss without the entropy regularization term. *We use a higher MSE threshold $m_2=0.001$ for No appr init model due to its difficulty in convergence.}
    \label{tab:ablation}
    \vspace{-1.5em}
\end{table}


\paragraph{Results} \tabref{tab:quantitative_all} shows the comparison to all baselines on unseen novel views. 
Overall, our method provides significantly better reconstruction quality over the all baselines, in all factorized regions using both synthetic and real world data. 
In real-world data, NeRF-time fails to reconstruct dynamic regions well and cannot provide reconstructions as good as NeRF in static regions. In contrast, our method can handle both regions well. NeRF-W can provide competitive reconstruction quality in pixel accuracy, but we still outperform it significantly in perceptual metric LPIPS. This demonstrates the benefits of using a factorized representation.  
\figref{fig:novel_view_comparison} shows the visual comparisons in synthetic and real world data respectively. Compared to all baselines, our method is the only the method that can reconstruct both static background and dynamic objects in details from novel views. 
We encourage the readers to watch the supplementary video which best demonstrates the perceptual quality of the rendering on novel spatial-temporal views.

\begin{figure}[t]
    \centering
    \tabcolsep 1.5pt
    \begin{tabular}{*5{c}}
        & No appr init & No online & No entropy & Full model \\
        \rotatebox[origin=c]{90}{Composition} &
        \raisebox{-.5\height}{\includegraphics[width=0.22\linewidth]{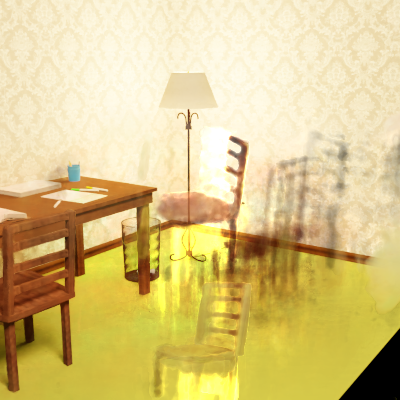}} &
        \raisebox{-.5\height}{\includegraphics[width=0.22\linewidth]{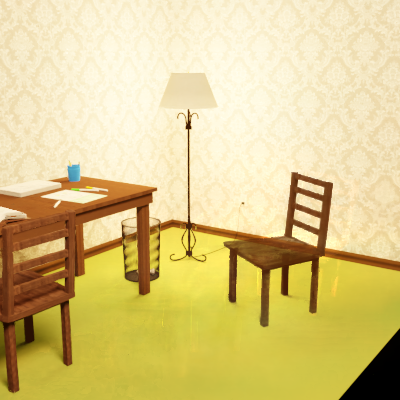}} &
        \raisebox{-.5\height}{\includegraphics[width=0.22\linewidth]{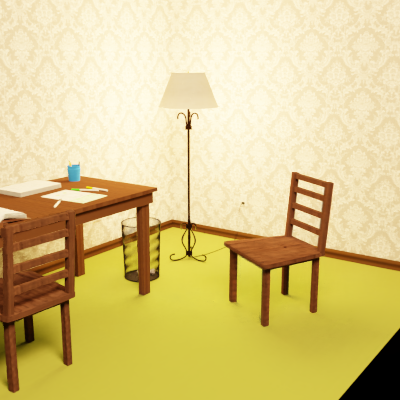}} & \raisebox{-.5\height}{\includegraphics[width=0.22\linewidth]{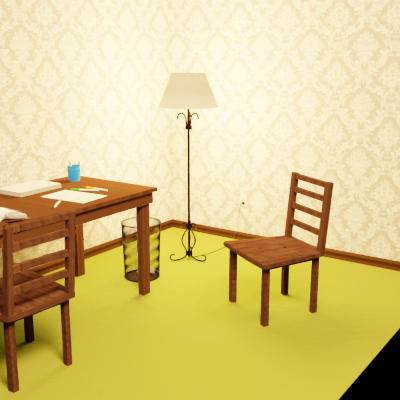}} \\
        \rotatebox[origin=c]{90}{Static} &
        \raisebox{-.5\height}{\includegraphics[width=0.22\linewidth]{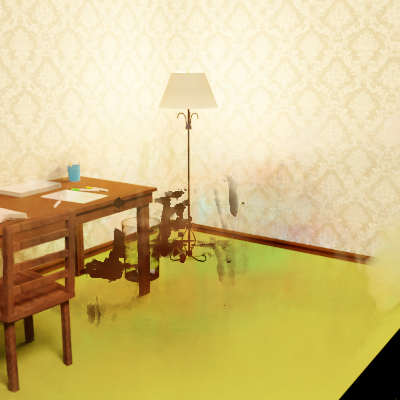}} &
        \raisebox{-.5\height}{\includegraphics[width=0.22\linewidth]{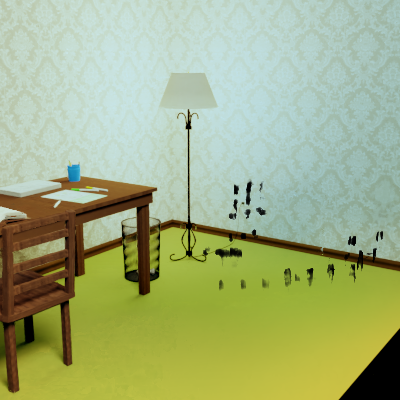}} &
        \raisebox{-.5\height}{\includegraphics[width=0.22\linewidth]{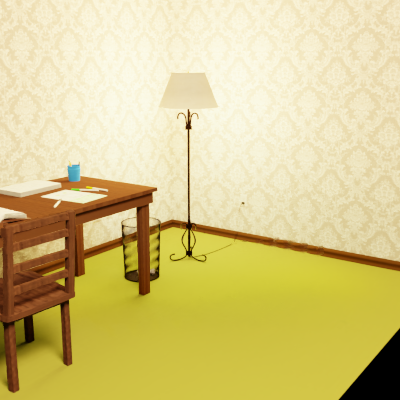}} & \raisebox{-.5\height}{\includegraphics[width=0.22\linewidth]{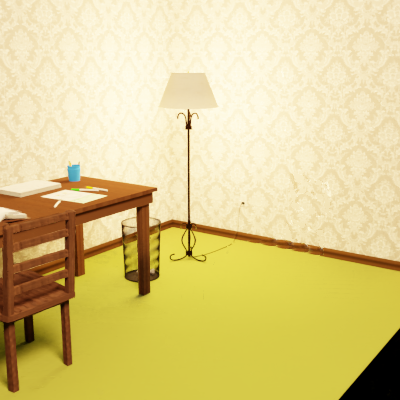}} \\
        \rotatebox[origin=c]{90}{Dynamic} &
        \raisebox{-.5\height}{\includegraphics[width=0.22\linewidth]{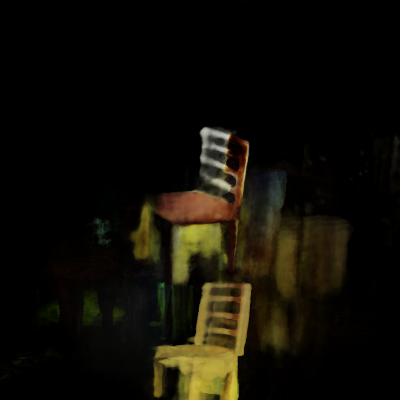}} &
        \raisebox{-.5\height}{\includegraphics[width=0.22\linewidth]{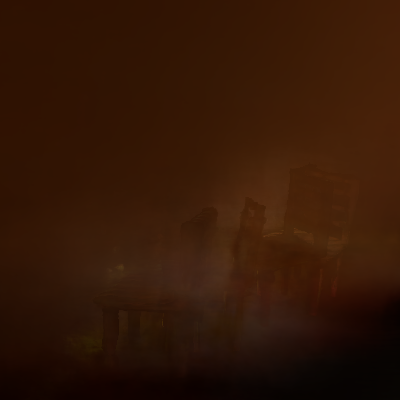}} &
        \raisebox{-.5\height}{\includegraphics[width=0.22\linewidth]{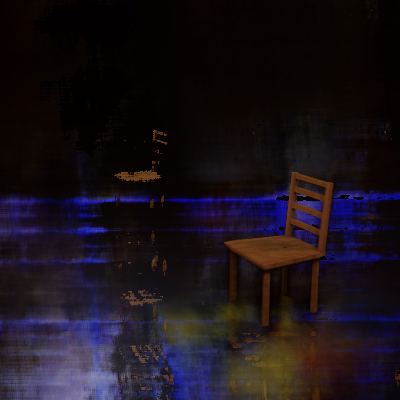}} & \raisebox{-.5\height}{\includegraphics[width=0.22\linewidth]{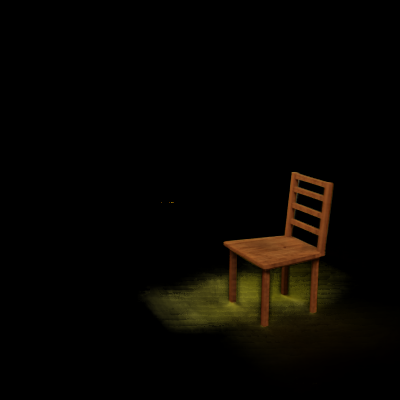}}
    \end{tabular}
    \caption{\small A visual comparison of the decomposition learned by our full model compared to ablated models.}
    \label{fig:ablation}
    \vspace{-1.5em}
\end{figure}

\paragraph{Ablation}
We perform an ablation study on our optimization strategies discussed in \secref{sec:optimization} using the \emph{lemp and desk} sequence. See quantitative results in \tabref{tab:ablation} and qualitative results in \figref{fig:ablation}. The results indicate our full model with regularization trained online performs the best. Our model performs poorly without the initialization stage, and also shows dramatically worse performance without the online training strategy. With additional entropy regularization over the volume density, we can further improve the results. \figref{fig:ablation} highlights the significant perceptual difference which cannot be properly seen in the quantitative numbers.

\subsection{Rendering on Animated Trajectories} 
\label{sec:anim}

Our factorized representation of motion and appearance allows STaR to synthesize novel views of animated trajectories of the dynamic object which have not been seen during training.  \figref{fig:animation} shows synthesized novel views of a trajectory dramatically different from the training trajectory in both synthetic and real world scenes. It is worth noting that no existing method we know of is able to synthesize motion so different from observed data and re-render it in a photorealistic fashion without any 3D ground truth or supervision, including NeRF-time and NeRF-W, which can only interpolate object poses from the observed trajectory.


\section{Conclusion}
\label{sec:conclusion}

Our method demonstrates a novel direction towards reconstructing dynamic scenes using only video observations. It should be noted here that this system is a proof of concept and has not completely solved the problem of fully decomposing dynamic scenes into their constituent parts.
First, we assume only one object in motion. Extending the model to multiple objects is trivial but estimating the number of moving objects when it is not known a priori is an interesting direction for further research.
Second, we cannot represent non-rigid motion in the presented model. This could probably be accomplished by combining our insights with the orthogonal work on deforming neural representations by Niemeyer et al. \cite{niemeyer:etal:cvpr19}.
Finally, we effectively remove geometric dynamism from both NeRF volumes by factorizing all motion into an explicit rigid transform, but we cannot do the same for appearance due to the mutual influence of each volume on the lighting conditions of the other.
This could be solved by further factoring the appearance into material and lighting conditions, but these explorations are very much outside the scope of this paper and we leave it for future work.

{\small
\bibliographystyle{ieee_fullname}
\bibliography{references}
}

\section*{Supplementary Material}
\renewcommand\thesection{\Alph{section}}
\renewcommand\theparagraph{\thesection.\Alph{paragraph}}
\setcounter{section}{0}

\begin{figure*}[h]
\centering
\subfloat[Reconstructed (left) and ground truth (right) chair in \textit{lamp and desk}]{
    \includegraphics[width=0.24\linewidth]{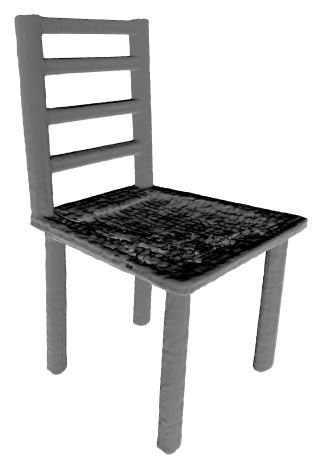}
    \includegraphics[width=0.24\linewidth]{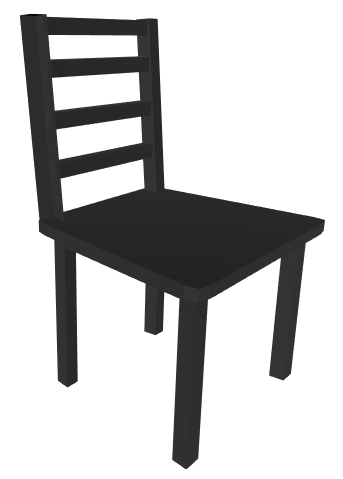}
}
\subfloat[Reconstructed (left) and ground truth (right) vase in \textit{kitchen table}]{
    \includegraphics[width=0.22\linewidth]{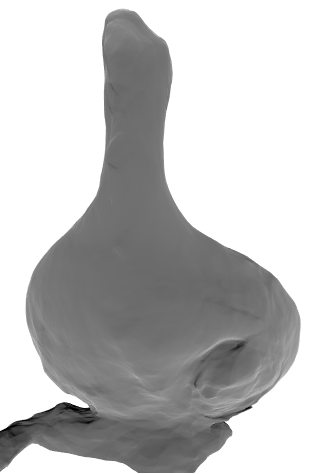}
    \includegraphics[width=0.26\linewidth]{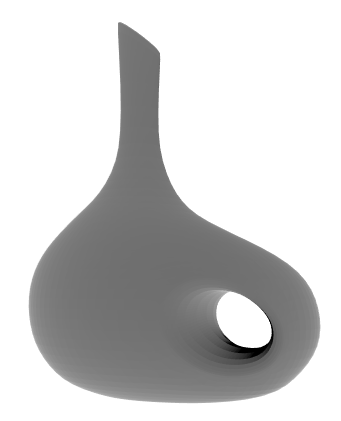}
}
\caption{Comparison of reconstructed mesh of the dynamic object and the ground truth.}
\label{fig:recon}
\end{figure*}

\section{Overview}
In this document, we provide technical details in support of our main paper. Below is a summary of the contents.
\begin{itemize}[leftmargin=1.2cm, noitemsep]
    \item[\secref{sec:video}:] Description of supplementary video;
    \item[\secref{sec:app}:] Mesh reconstruction and 6D pose tracking results;
    \item[\secref{sec:arch}:] MLP architecture and volume rendering details;
    \item[\secref{sec:data}:] Synthetic and real-world data preparation;
    \item[\secref{sec:grad}:] Derivation of $\mathrm{SE}(3)$ pose Jacobian.
\end{itemize}

\section{Video}
\label{sec:video}
We encourage the reader to watch our supplementary video at \url{https://wentaoyuan.github.io/star}, where we visualize the following results.
\begin{itemize}
    \item We first show a comparison of STaR against NeRF \cite{mildenhall:etal:arXiv20}, NeRF-time and NeRF-W \cite{martin:etal:arXiv20} on the rendering of novel spatial-temporal views on the \textit{lamp and desk} and \textit{kitchen table} sequences. The rendered videos are 20x slow motion of the training videos from a continuously varying camera view unseen during training, where STaR achieves superior perceptual quality compared to the baselines.
    \item Then, we visualize the decomposition of static and dynamic components learned by STaR on the \textit{moving banana} sequence by showing how static background and dynamic foreground can be seamlessly removed during spatial-temporal novel view rendering. Similarly, the rendered video is a 20x slow motion of the training videos from a continuously varying camera view unseen during training. We also visualize the disparity map rendered by STaR.
    \item Finally, we show results of photorealistic animation of unseen object motion in \textit{lamp and desk} and \textit{moving banana}. Specifically, we compose the static and dynamic NeRFs using a set of poses significantly different from the poses seen during training (see \figref{fig:animation} for a visualization of the trajectories) and rendered the composed NeRF from a continuously varying camera view unseen during training. Remarkably, without any prior knowledge about the scene geometry of the object motion, STaR is able to learn a factored representation that allows it to photorealistically synthesize novel spatial-temporal views of novel object motion, which no existing method can do.
\end{itemize}
  
\section{Additional Applications} \label{sec:app}

\paragraph{Mesh Reconstruction}
The separation of static and dynamic components learned by STaR allows us to reconstruct a 3D mesh of the unknown dynamic object. Specifically, we can query the dynamic NeRF using a dense 3D grid over a training camera's view frustum, then threshold the density outputs (i.e. setting $\sigma^D=0$ if $\sigma^D<\sigma_{\min}$) and run marching cubes to obtain a 3D mesh. \figref{fig:recon} visualizes the reconstructed meshes of the dynamic objects compared to the ground truth in the two synthetic videos used in our paper, \textit{lamp and desk} and \textit{kitchen table}.

We also use MeshLab to compute the mean Hausdorff distance between the reconstructed mesh and the ground truth. We report the distances in \tabref{tab:recon} as percentage of the ground truth mesh's bounding box diagonal. We use voxel size 0.002 for \textit{lamp and desk}, voxel size 0.0001 for \textit{kitchen table}, and density threshold 50 for both scenes. The reconstructed meshes are post-processed by excluding everything outside of a manually specified 3D bounding box and aligned with the ground truth meshes using ICP.

\begin{table}[h]
    \centering
    \begin{tabular}{c|cc}
        \toprule
         & Lamp and desk & Kitchen table \\ \midrule
        Hausdorff distance & 0.55\% & 3.59\% \\
        \bottomrule
    \end{tabular}
    \caption{Hausdorff distance between the reconstructed and ground truth mesh as percentage of the ground truth mesh's bounding box diagonal.}
    \label{tab:recon}
\end{table}

\paragraph{6DoF Pose Tracking}
In addition to reconstructing geometry and appearance, STaR also outputs the relative 6D pose between the static and dynamic volume. We can use the output poses to accurately track the relative motion of the dynamic object. In \tabref{tab:track}, we report error in the relative pose difference of the dynamic object between neighboring key frames estimated by STaR compared against the ground truth. The rotation error is computed in degree and the translation error is computed as percentage of the diagonal of the object's 3D bounding box.

\begin{table}[h]
    \centering
    \begin{tabular}{c|cc}
        \toprule
         & Lamp and Desk & Kitchen table \\ \midrule
        Rotation error & 0.502 & 3.198 \\
        Translation error & 0.76\% & 3.60\% \\
        \bottomrule
    \end{tabular}
    \caption{Rotation and translation error of the dynamic object's relative motion between key frames estimated by STaR. Translation error is computed as percentage of the object's bounding box diagonal.}
    \label{tab:track}
\end{table}

\begin{figure*}[h]
\centering
\subfloat[STaR (NeRF)]{
    \includegraphics[width=0.32\linewidth]{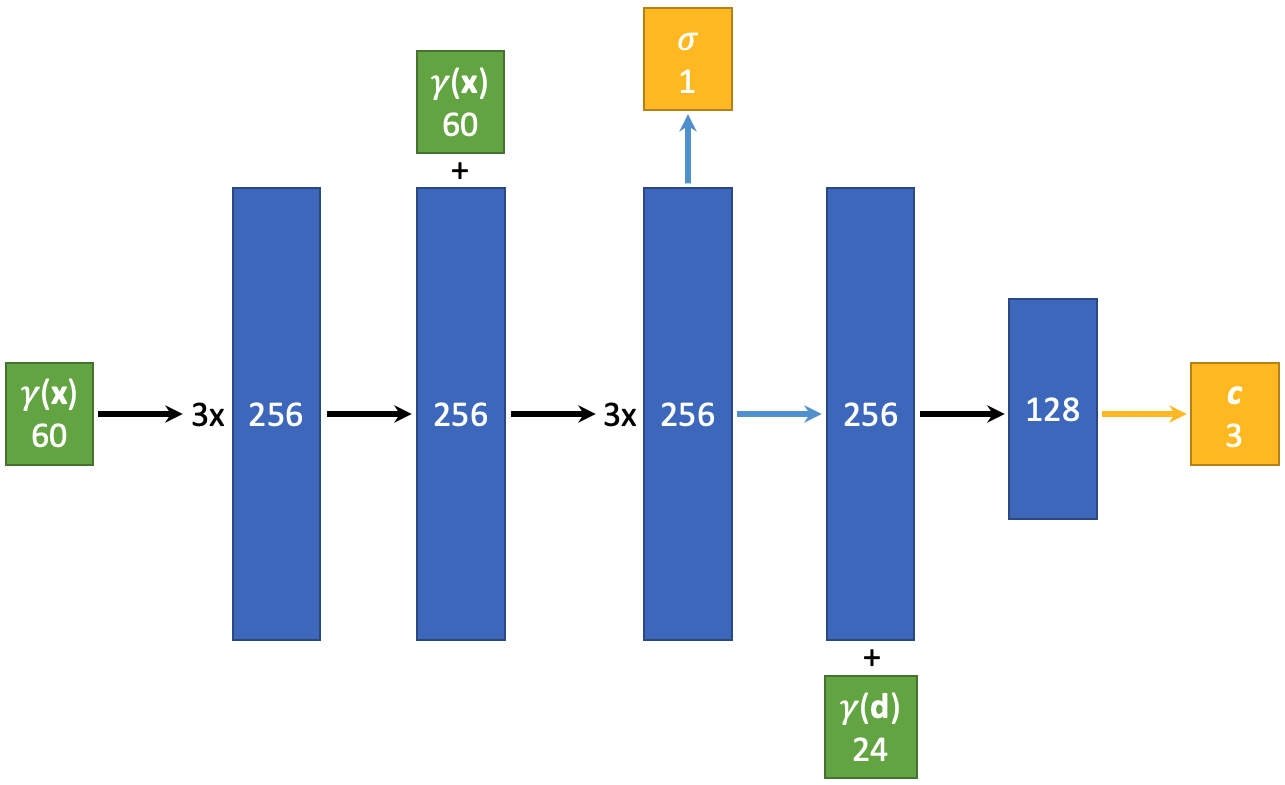}
} \hfill
\subfloat[NeRF-time]{
    \includegraphics[width=0.32\linewidth]{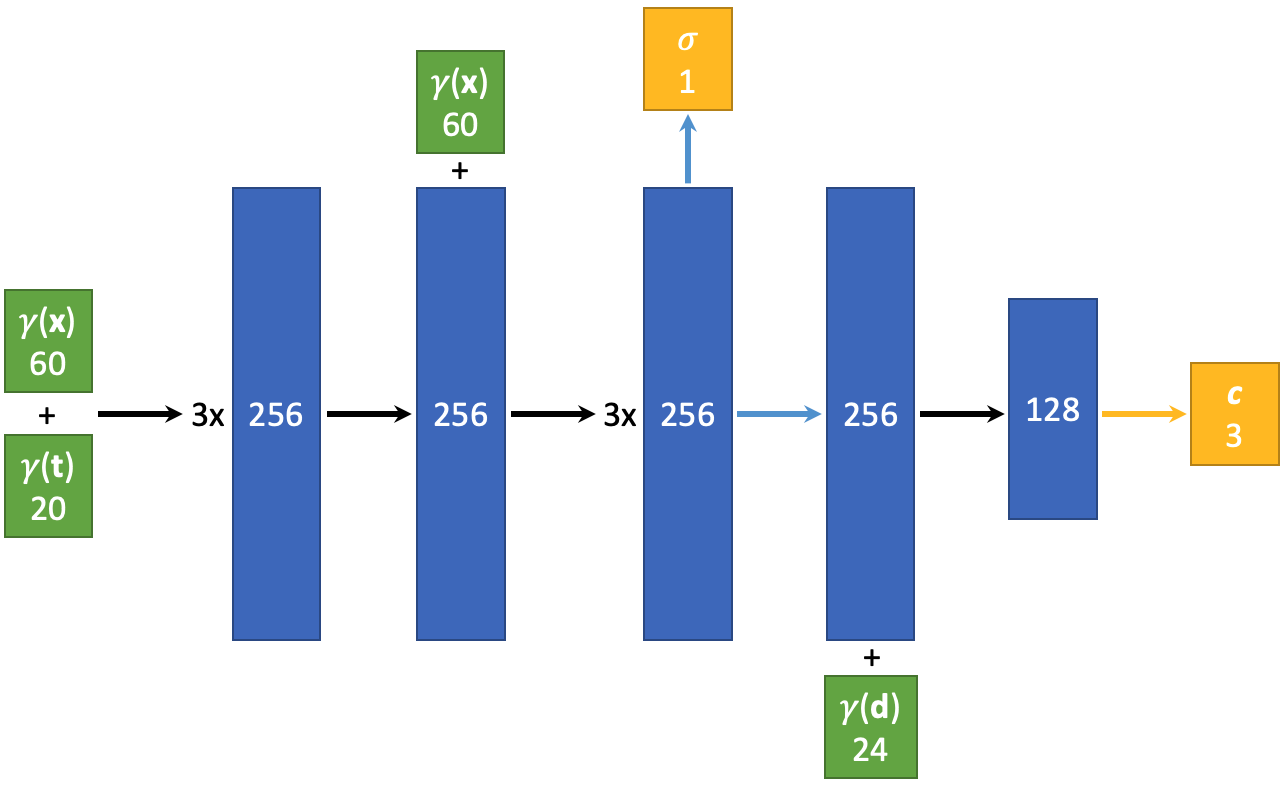}
} \hfill
\subfloat[NeRF-W (NeRF-U)]{
    \includegraphics[width=0.32\linewidth]{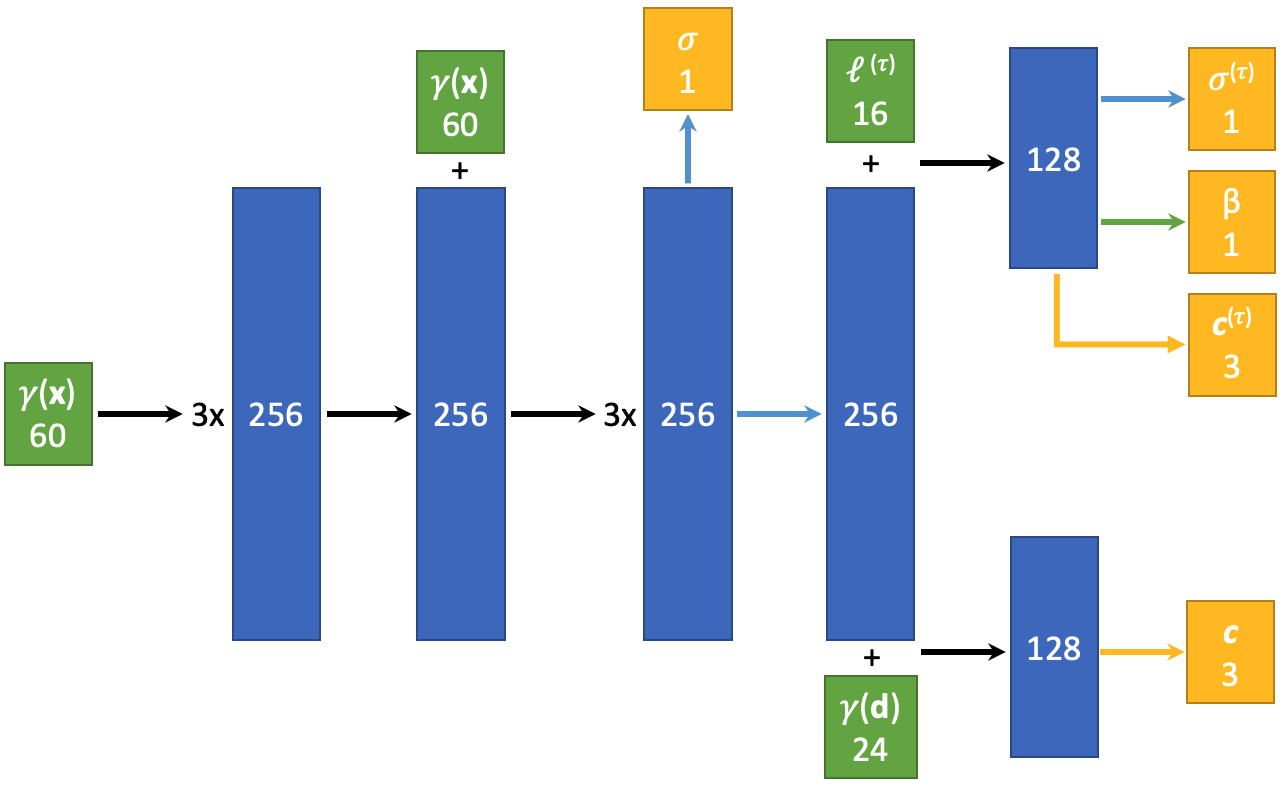}
}
\caption{Architecture of MLPs used in STaR (NeRF), NeRF-time and NeRF-W. \textcolor{green}{Input}, \textcolor{blue}{network layers} and \textcolor{yellow}{outputs} are marked in green, blue and yellow respectively, with their dimensions labeled beneath. Blue, black, yellow and green arrows denote linear transformations with \textcolor{blue}{no activation}, ReLU activation, \textcolor{yellow}{sigmoid activation} and \textcolor{green}{softplus} activation respectively and $+$ denotes concatenation. $\gamma$ denotes positional encoding and $\mathbf{x},\mathbf{d},\sigma,\mathbf{c}$ denotes 3D location, viewing direction, volume density and radiance. $l^{(\tau)}$ is the latent code taken by NeRF-W to generate transient density $\sigma^{(\tau)}$, color $\mathbf{c}^{(\tau)}$ and uncertainty $\beta$.}
\label{fig:mlp}
\end{figure*}

\section{Implementation Details} \label{sec:arch}

\paragraph{MLP Architecture}
\figref{fig:mlp} shows the architecture of MLPs used by STaR (NeRF), NeRF-time and NeRF-W respectively. STaR uses the same MLP architecture as NeRF for both static volume $\mlp^S$ and dynamic volume $\mlp^D$. NeRF-time shares the same MLP architecture except for using positional-encoded time as additional input. In practice, the time parameter before positional encoding is the frame index linearly projected on to the interval $[-1,1]$. NeRF-W (more precisely, its variant NeRF-U since we don't use appearance embedding) uses the same MLP architecture as NeRF for the coarse network, but its fine network takes an additional 16-dimensional latent code and outputs transient density, color and uncertainty in addition to static density and color. Please refer to \cite{martin:etal:arXiv20} for more details about the architecture of NeRF-W.

\paragraph{Volume Rendering}
For STaR and all baselines, we use 64 stratified samples per ray for the coarse network and additional 64 importance samples (in total 128 samples) for the fine network. Following \cite{mildenhall:etal:arXiv20}, we add small Gaussian noise to the density outputs during appearance initialization but turn it off during online training. We adjust the absolute scale of the camera's view frustum so that it roughly lies within the cube $[-1,1]^3$. For synthetic data, this can be done by scaling the rendered content. For real data, we translate the camera poses so that the world coordinate center aligns with the center of the average camera's view frustum. Then we scale the camera poses' translation component by half of the distance from the near bound to the far bound.

\section{Data Preparation} \label{sec:data}

\paragraph{Synthetic Data Generation}
The synthetic video sequences are rendered with Blender Cycles rendering engine using photorealistic assets created by professional designers on Blend Swap. The 8 training cameras are arranged in a $2\times4$ array, focusing on the same target point in the scene. The evaluation camera is also looking at the same direction from a similar distance but its location is different from the training cameras. The camera poses remain fixed throughout the video.

\paragraph{Real World Data Capture}
The real-world video sequence is captured using a 20-camera rig. The cameras are arranged in a $2\times10$ array. We discard 3 cameras that are not synced with the others, use 16 cameras for training and 1 remaining camera for evaluation. The camera poses are also fixed and can be obtained by running COLMAP's structure from motion pipeline on images from the first frame. The original image resolution is $2704\times2028$ and is downsampled to $676\times507$ for training and evaluation. 

\section{Jacobian for Pose Gradient} \label{sec:grad}
Let $\mathbf{p}\in\R^3$ be a 3D point and $T\in\mathrm{SE}(3)$ be a pose with associated transformation matrix
\begin{equation}
    T = \begin{bmatrix}
    \mathbf{r}_1 & \mathbf{r}_2 & \mathbf{r}_3 & \mathbf{t} \\
    \mathbf{0} & \mathbf{0} & \mathbf{0} & 1
    \end{bmatrix}
\end{equation}
where $T[\mathbf{p}]$ denotes the transformation of $\mathbf{p}$ with respect to $T$.
Let $\epsilon\in\se{3}$ be a local perturbation on the $\mathrm{SE}(3)$ manifold. We are interested in the derivative of $\exp(\epsilon)T[\mathbf{p}]$ with respect to $\epsilon$ at $\epsilon=0$:
\begin{align}
    \left.\frac{\partial\exp(\epsilon)T[\mathbf{p}]}{\partial\epsilon}\right|_{\epsilon=0} &= \left.\frac{\partial S[\mathbf{p}]}{\partial S}\right|_{S=\exp(\epsilon)T=T} \left.\frac{\partial\exp(\epsilon)[T]}{\partial\epsilon}\right|_{\epsilon=0} \\
    &= \left(\begin{bmatrix}\mathbf{p}^\top & 1\end{bmatrix}\otimes \mathbf{I}_3\right)
    \begin{bmatrix}
    \mathbf{0}_{3\times3} & -\mathbf{r}_1^\wedge \\ \mathbf{0}_{3\times3} & -\mathbf{r}_2^\wedge \\
    \mathbf{0}_{3\times3} & -\mathbf{r}_3^\wedge \\ \mathbf{I}_3 & -\mathbf{t}^\wedge
    \end{bmatrix} \\
    &= \begin{bmatrix} \mathbf{I}_3 & -(T[\mathbf{p}])^\wedge \end{bmatrix}
\end{align}
where $\otimes$ denotes Kronecker product.
The result is a $3\times 6$ Jacobian matrix, where $\mathbf{x}^\wedge$ is the cross product matrix
\begin{equation}
    \mathbf{x}^\wedge = \begin{bmatrix}
        0 & -x_3 & x_2 \\
        x_3 & 0 & -x_1 \\
        -x_2 & x_1 & 0
    \end{bmatrix}
\end{equation}
We encourage the reader to read \cite{blanco2010tutorial} for more details about the on-manifold optimization of $\mathrm{SE}(3)$ transformations.



\end{document}